\documentclass{article}
\usepackage{amsmath,amsfonts,bm}
\usepackage{gensymb}

\def\eqref#1{equation~\ref{#1}}
\def\1{\bm{1}}

\DeclareMathAlphabet{\mathsfit}{\encodingdefault}{\sfdefault}{m}{sl}
\SetMathAlphabet{\mathsfit}{bold}{\encodingdefault}{\sfdefault}{bx}{n}

\DeclareMathOperator*{\argmax}{arg\,max}

\usepackage{fullpage}
\usepackage{color,graphicx}
\usepackage{dcolumn}
\newcolumntype{d}[1]{D{.}{.}{#1}}
\title{Computational Ceramicology}
\author{Barak Itkin \and Lior Wolf\and Nachum Dershowitz}
\begin{document}

\maketitle
\begin{abstract}

Field archeologists are called upon  to identify potsherds, for which purpose they rely on their experience and on reference works. We have developed two complementary machine-learning tools to propose identifications based on images captured on site. One method relies on the shape of the fracture outline of a sherd; the other  is based on decorative features.
For the outline-identification tool, a novel deep-learning architecture was employed, one that integrates shape information from points along the inner and outer surfaces.
The decoration classifier is based on relatively standard architectures used in image recognition. In both cases, training the classifiers required tackling challenges that arise when working with real-world archeological data: paucity of labeled data; extreme imbalance between instances of the different categories; and the need to avoid neglecting rare classes and to take note of minute distinguishing features of some classes. The scarcity of training data was overcome by using synthetically-produced virtual potsherds and by employing multiple data-augmentation techniques. A novel form of training loss allowed us to overcome the problems caused by under-populated classes and non-homogeneous distribution of discriminative features.
\end{abstract}
\section{Introduction}

Pottery is the most common type of excavated artifact, and its identification permits the understanding of the chronology, the function, and the importance of an archeological site. This identification is based on the archeologist's domain knowledge and is usually done by matching potsherds to exemplars  in catalogues of semi-standardized archeological typologies. These catalogues contain, for each type, a standardized sketch of the complete vessel and sometimes a few photos of excavated samples. 

The set of tools we develop addresses two scenarios: (i) In the more common case, the pottery is undecorated, either because it was manufactured that way or because its decorations have been lost due to the ravages of time. In this case, the identification relies on the geometry of the sherd. (ii) If visual patterns, such as colors and decorations, are preserved, classification is usually based on those, since they can be more diagnostic than the shape of the sherd. 

\begin{figure*}[t]
\centering
  \begin{tabular}{c@{\hskip 0.05\linewidth}c@{\hskip 0.05\linewidth}c}
\includegraphics[width=0.20\linewidth]{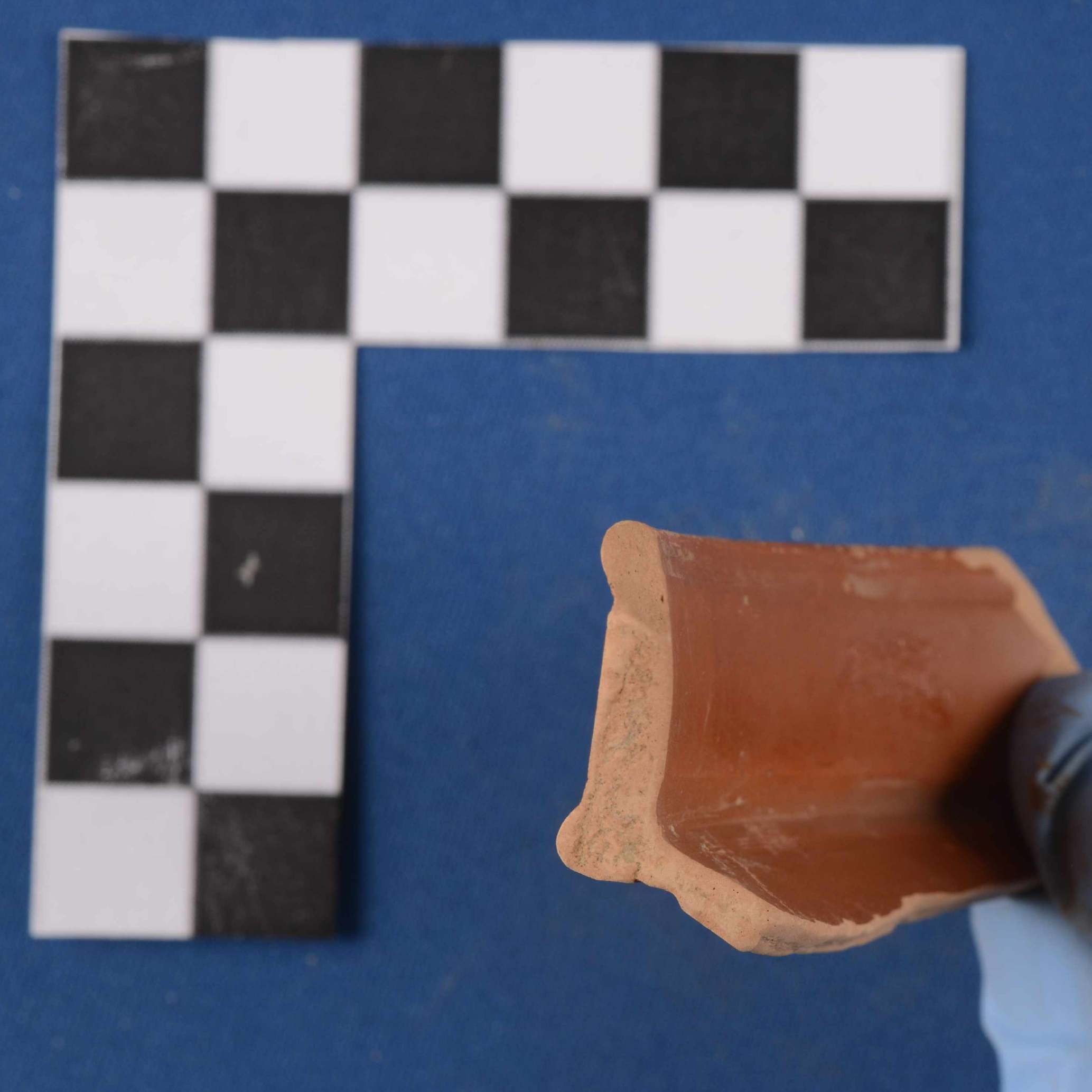}&
\includegraphics[width=0.20\linewidth]{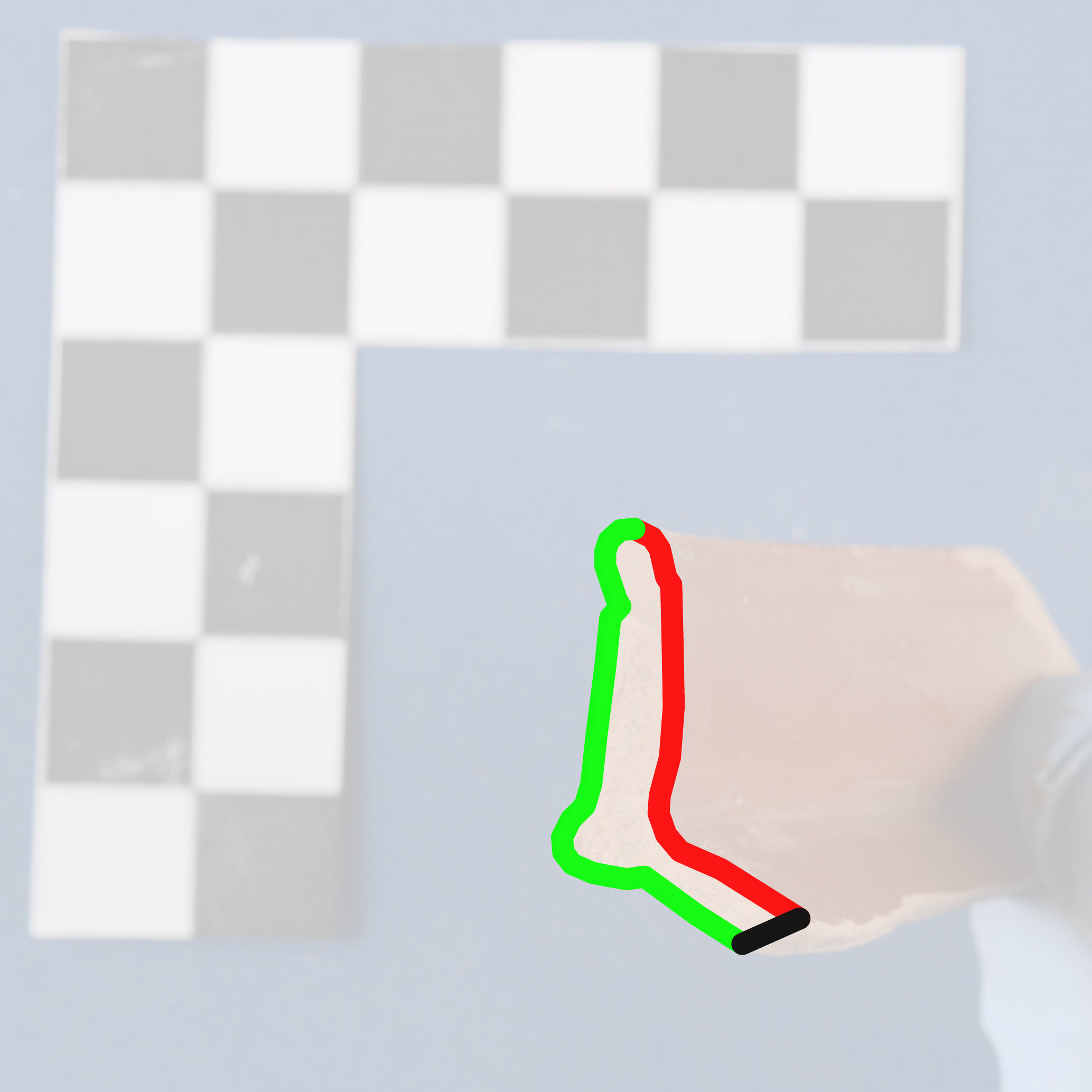}&
\includegraphics[width=0.20\linewidth]{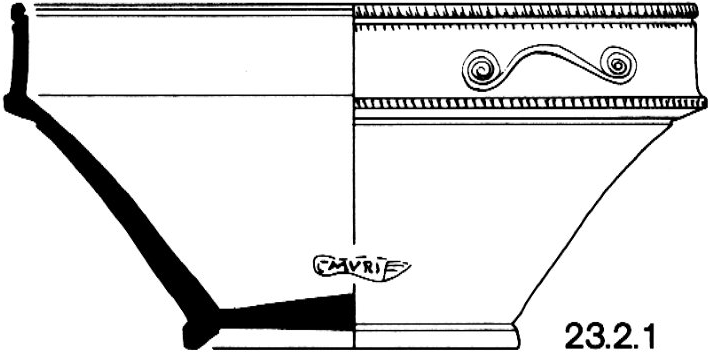}\\
(a) & (b) & (c)\\
~\\
\includegraphics[width=0.20\linewidth]{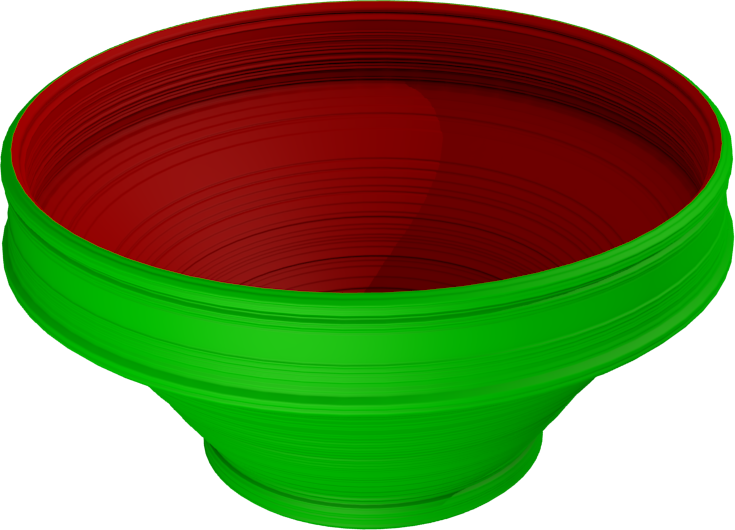}&
\includegraphics[width=0.20\linewidth]{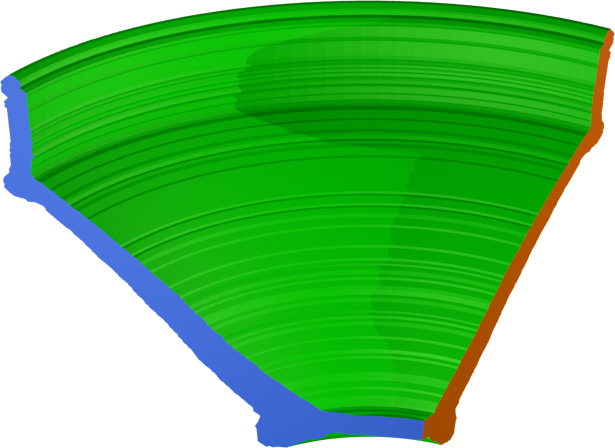}&
\includegraphics[width=0.20\linewidth]{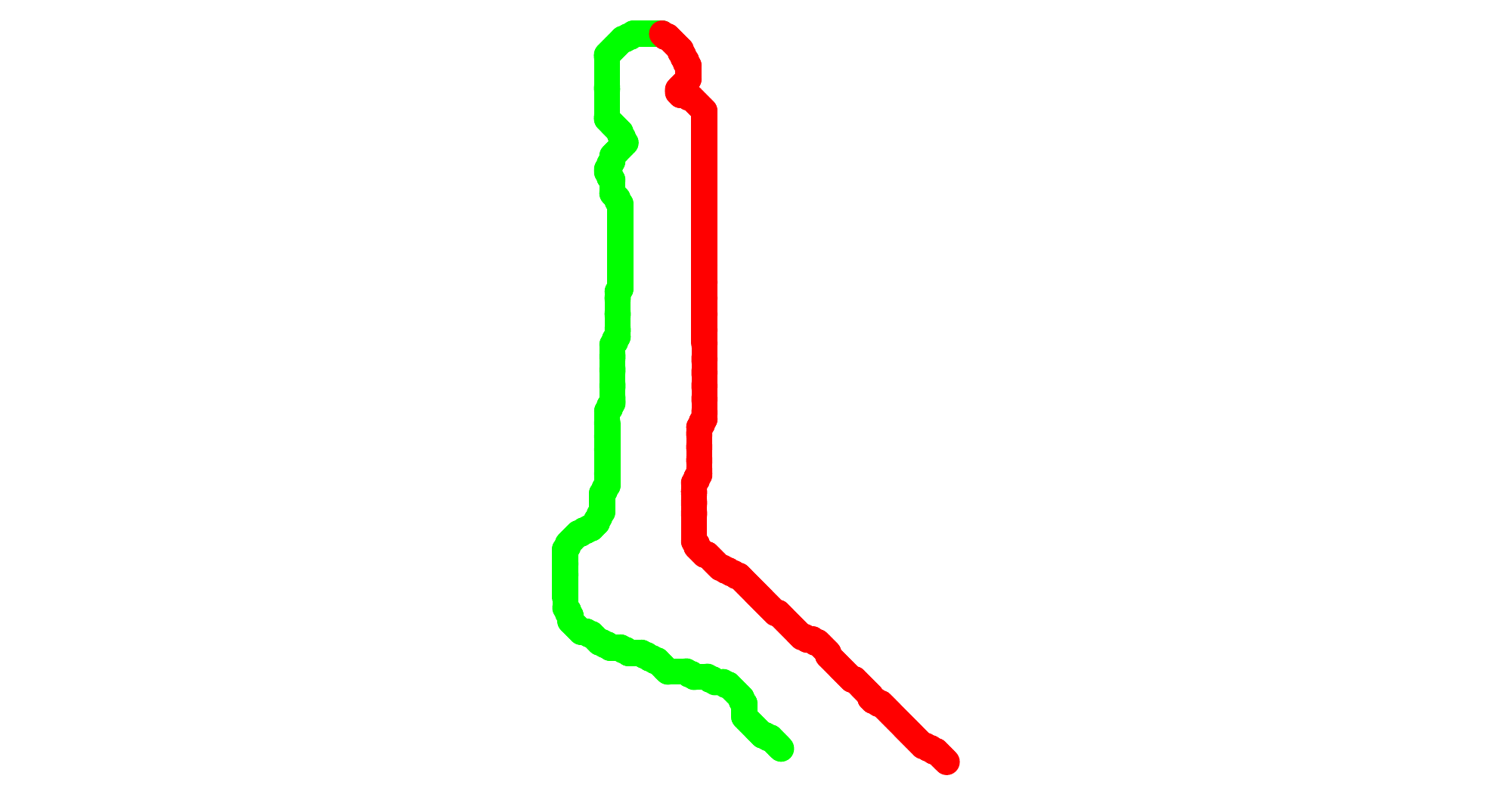}\\
(d) & (e) & (f)\\
\end{tabular}
\caption{An illustration of the archeological data. (a) An image of a sherd, positioned to show the fracture surface, with a reference scale ruler in the background. (b) A traced fracture outline, overlaid over the source image. Green is the outer profile; red is the inner profile; pink is for  break lines that are ignored by the algorithm. (c) An archeological sketch as it appears in a catalog. One or more sketches define a class of pottery.  (d) A 3D computer graphics vessel obtained by rotating the catalog sketch. (e) A synthetic sherd obtained by breaking the 3D vessel. (f) A fracture outline obtained directly from the sketch, without the 3D reconstruction and shattering process.}
\label{fig:leftsherdrightsketchs}
\end{figure*}

\paragraph{Shape-based identification.} 
Since our research is aimed toward aiding archeologists in the field, we forgo complex attempts to extract 3D geometry and rely on the 2D outline of the fracture surface of the sherd as the source of shape information.  We tackle the task of classifying the outline of a potsherd based on a single image of it, as depicted in Figure~\ref{fig:leftsherdrightsketchs}(a). After marking the outline in a semi-automatic way and determining the scale using a ruler (Figure~\ref{fig:leftsherdrightsketchs}(b)), our AI-powered mobile app provides the identification in the form of a list of archeological types, ranked  by their relevance to the pictured potsherd. 

A major challenging in training the AI toold is that one cannot obtain enough training samples that are similar to those used in test; only a handful of available sherds per class have been digitized. Furthermore, even if all extant sherds were to be digitized, the variability in the dataset would still cover only a small part of the space of possible sherds. Instead, we define each class by one or more 2D sketches of the profile of the complete vessel; see Figure~\ref{fig:leftsherdrightsketchs}(c). Note that the sketch describes the geometry of the profile of the entire vessel, and the excavated sherd is a relatively small piece of the original, containing very limited information regarding the  shape as a whole.

The outline of the fracture is a consequence of both the geometry of the pottery, and of the random breakage process. On the dataset side, we  reconstruct the 3D pottery by rotating the profile of the vessel (Figure~\ref{fig:leftsherdrightsketchs}(d)) and shatter it in order to derive synthetic sherds (Figure~\ref{fig:leftsherdrightsketchs}(e)). We propose a way to avoid the associated computation overhead of 3D reconstruction and instead, obtain the synthetic fracture surface (Figure~\ref{fig:leftsherdrightsketchs}(f)) directly, and even on-the-fly during training. 

To identify outlines, we train a network that supports the unique characteristics of archeological outlines, including the need to separate between the inner outline and the outer outline of the sherd, the importance of the order of the points along the outline, the inherent noise in the tracing process, and the need to overcome sub-optimal data acquisition processes.  

\paragraph{Decoration identification.} 
The case in which the color information on the face of the artifact is informative, is much better addressed in the current computer vision literature. In this case, we employ a commonly used {\em transfer-learning} technique in which a neural network that was pre-trained to perform visual identification is adopted to the task at hand, using the relatively small archeological training dataset.

In both the shape and the decoration methods, we are required to overcome a large set of compounding challenges. These include:
(1) the lack of real-world data to train on (shape) or a small one (decorations);
(2) a partial view of the object that is obtained by a random breakage process, which presents large variability;
(3) a large portion of the sherds, among both the synthetic training samples and the captured test samples, are almost completely non-informative; 
(4) very similar classes, making the distinction more challenging and also causing ambiguity in the ground truth classification of the test data;
(5) a noisy acquisition process: an error prone process for extracting the outline and obtaining scale from the real images (shape), variability in illumination (decorations).

In addition, in order to be used by experts, there is an acute need to optimize to fit considerations beyond accuracy. For example, most neural network losses would be prone to sacrificing challenging classes in order to improve the average accuracy across all classes; However, a reference tool brings the most value when the identification is less obvious. To tackle the heterogeneous and unbalanced nature of the data, we train using a novel weighting technique that considers both the error of each ground truth class and false positives in each class. 

Our results demonstrate a relatively high recognition rate in the face of these challenges. To ensure the validity of our results, development was carried out in two phases. In the first, we developed the method on one dataset of potsherds of one specific family; in the second, the same method, with exactly the same pipeline and (hyper-) parameters, was applied to three new datasets. With our Phase I dataset, out of 65 different classes, we are able to identify---based on images of sherds captured with a dedicated mobile app---almost 74\% of the sherds within the top-10 results. With three additional datasets that were received after the completion of our research phase, without any tweaking of the pipeline, we reached 81\%, 68\%, and 60\% top-10 accuracy for 65, 98, and 94 classes, respectively. Thus, our network may serve as the basis of a reliable reference tool for the use of archeologists in the field, one that significantly narrows down the list of relevant classes to be considered for each sherd. 

\section{Related Work}

\subsection{Sim2Real}
In our work, we bridge the semantic gap between sketches and excavated potsherds by training on synthetic data. A common term that is used in such cases is sim2real (simulation to real world). The most popular use case is to overcome the sample complexity of current reinforcement learning (RL) methods, when learning complex policies. Since it has been demonstrated that policies are learnable in simulation worlds~\cite{peng2016terrain,peng2017deeploco}, much of the current effort in RL focuses on learning in simulated environments  (\cite{andrychowicz2018learning,tan2018sim,peng2018sim} to cite a few examples).

Computer graphics animations and images are extensively used to train and evaluate deep optical flow networks~\cite{fischer2015flownet}, detect text~\cite{gupta2016synthetic} or object instances~\cite{hodan2019photorealistic} in images, understand indoor scenes~\cite{handa2016understanding}, and estimate the pose of humans~\cite{varol17_surreal} or objects~\cite{tremblay2018deep}, among other tasks. In fact, the same computer game data can be used, for example, to solve many tasks, including optical flow, semantic instance segmentation, object detection and tracking, object-level 3D scene layout,  visual odometry~\cite{DBLP:conf/iccv/RichterHK17}, or---going back to the RL applications---driving simulators can train a real-world robotic driver~\cite{bewley2018learning}. 

\subsection{Automated Pottery Classification}
In the absence of organic material to allow for carbon dating, pottery classification provides an indispensable tool for dating excavated content. 
Much of the work on automated identification of sherds is based on 3D scanning or multi-view reconstruction technologies~\cite{Kampel06,Karasik10,Calin12,Barreau14}. However, the adoption of such methods is very limited due to the challenges of 3D acquisition in the field. In addition, the challenges of analyzing 3D shapes have only been partially solved. 

The automatic analysis of profiles of potsherds has been studied using classical computer vision methods, such as the Hough transform~\cite{Durham95}, various morphological features~\cite{Karasik11,Lucena14}, and curvature descriptions~\cite{Gilboa04}. None of these systems is robust enough to be applied automatically on a varied set of excavated sherds and much of it was only applied to complete profiles. As mentioned above, appearance-based methods~\cite{Makridis13,Piccoli13} are not appropriate for the type of sherds considered in this work.

\subsection{Generating Potsherds from Profile Drawings}
The problem of reconstruction from line drawing or sketches is a classical problem (e.g.\@ \cite{Malik87, Tian09, Wang09, Xu14}). Architects, similar to archeologists, have used semi-structure sketches, which can aid in reconstruction; see~\cite{Yin09}. 

Our dataset-side pipeline of extracting synthetic sherds based on catalogical sketches (not including any of the subsequent learning techniques) solves a similar task to the one presented by~\cite{icdar}. In that work, it was suggested to reconstruct the 3D model of a class by rotating the inner and outer profiles around a rotation axis. To make it computationally feasible to generate a model from an outline with thousands of points, they use outline-simplification algorithms to narrow down the number of points. Such simplifications are detrimental, however, since some   pottery present delicate details, and the discriminative parts are sometimes only 1--2 cm long. Since the size of the rotational model is quadratic in the number of profile points (typically several thousand), without simplification, the generation of millions of training samples is infeasible.

\subsection{PointNets and Similar Architectures}
The architecture of our classifier relates to an emerging body of work,  encoding inputs that are given as sets~\cite{qi2017pointnet,NIPS2017_6931}. It is similar to PointNet~\cite{qi2017pointnet} in that it employs pooling in order to obtain a representation that is invariant to the order of the elements, following a local computation at each element. It was previously shown by~\cite{qi2017pointnet,NIPS2017_6931} that, under mild conditions, such pooling is the only way to achieve this invariance. 

Other recent works in the area of shape classification include PointNet++~\cite{qi2017pointnetplusplus}, which employs local spatial relations, and PointCNN \cite{hua2018pointwise}, which applies spatial information, in order to group the points prior to aligning them spatially to a grid where a convolution can be applied. While  previous work is mostly focused on the identification of 3D point clouds, we encode a 2D outline, and benefit from information that arises from the order of the points along the outline. In addition, projected profiles of 3D objects have an inner profile and an outer profile, and the separation between the two carries valuable information. 
	
\subsection{Data Reweighting Schemes}
Boosting techniques often iteratively weight harder samples, which are misclassified during training, more than other samples~\cite{freund1997decision}. In detection, such hard negatives are of great importance~\cite{viola2001rapid}, and, as suggested more recently by the focal loss method of~\cite{lin2018focal}, assigning different weights to the loss of different examples can significantly improve the result of the training. Another common cause for introducing weights into the loss function is class imbalance of the available samples, and it is common to assign higher weights to less frequent classes. 

The reweighting scheme that we propose here addresses both the difficulty of correctly classifying a sample from a given class, as well as the frequency of the current classification of a sample. While the classification difficulty component is somewhat similar to other methods, the other component is, as far as we can ascertain, completely novel. 

Our reweighting scheme improves, not just the top-1 result, but also the top-$k$ and is, accordingly, related to recent methods in this field. While it has been proven that the softmax-based cross entropy loss is optimal for every $k$ under  i.i.d.\@ sampling and infinite data assumptions~\cite{lapin2018analysis}, these assumptions do not hold in our case in which there is a significant domain shift between train and test data. Recently, a method was proposed to overcome the infinite data assumption, by employing a smoothed variant of a novel top-$k$ SVM formulation~\cite{berrada2018smooth}. This method, however, does not account for domain shift. Unsupervised domain adaption~\cite{DBLP:conf/colt/MansourMR09} techniques are designed to overcome such a shift. However, these methods, including the recent adversarial-training based ones following~\cite{Ganin:2016:DTN:2946645.2946704}, are not suitable for our case either since we do not possess a significant unsupervised set from the target domain. Since our method is orthogonal to the type of loss used and to the use of additional loss terms on the intermediate features, it can be used in concert with those methods.

\section{The Shape-Based Tool}

\subsection{Synthetic  Training Data}
\label{sec:doc2sherd}

Generating high quality data, with as much similarity to  real data as possible, is critically important for training an automatic sim2real classification model. To generate synthetic training data using the sketches extracted from the catalogs, our process follows the steps described next.

\begin{figure}[tb]
\centering
\begin{tabular}{cc}
\includegraphics[width=0.48\linewidth]{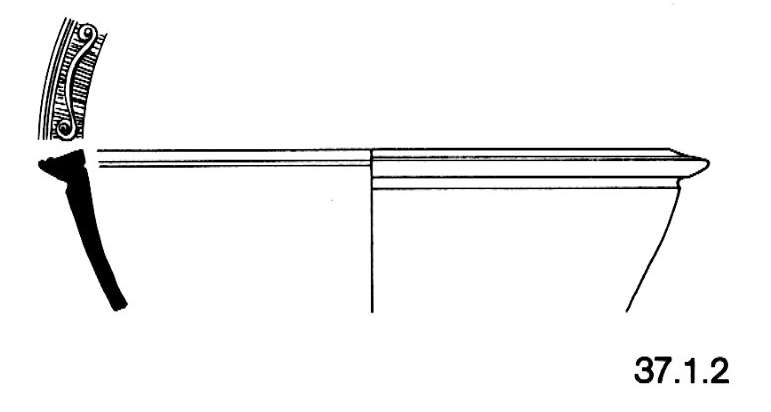}&
\includegraphics[width=0.48\linewidth]{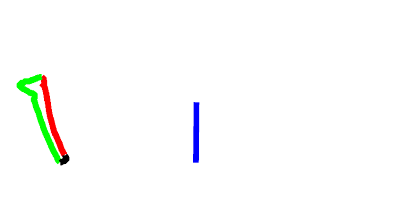}\\
(a) & (b) \\
\end{tabular}\caption{An example of a partial profile. (a) Drawing of a profile describing only the rim of the vessel. (b) The extracted profile and rotation axis, with ``missing'' areas manually annotated in black.}
\label{fig:partial_profile}
\end{figure}

\begin{figure*}[tb]
\centering
\begin{tabular}{cccc}
\includegraphics[width=0.212402431213\linewidth]{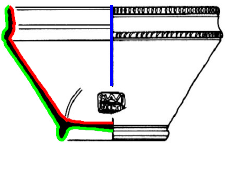}&
\includegraphics[width=0.212402431213\linewidth]{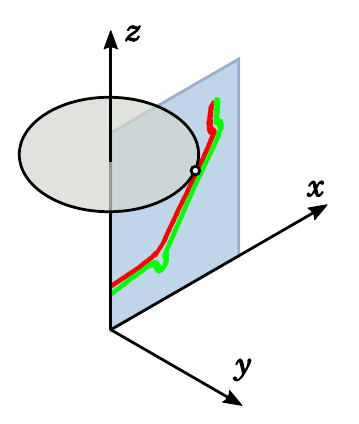}&
\includegraphics[width=0.212402431213\linewidth]{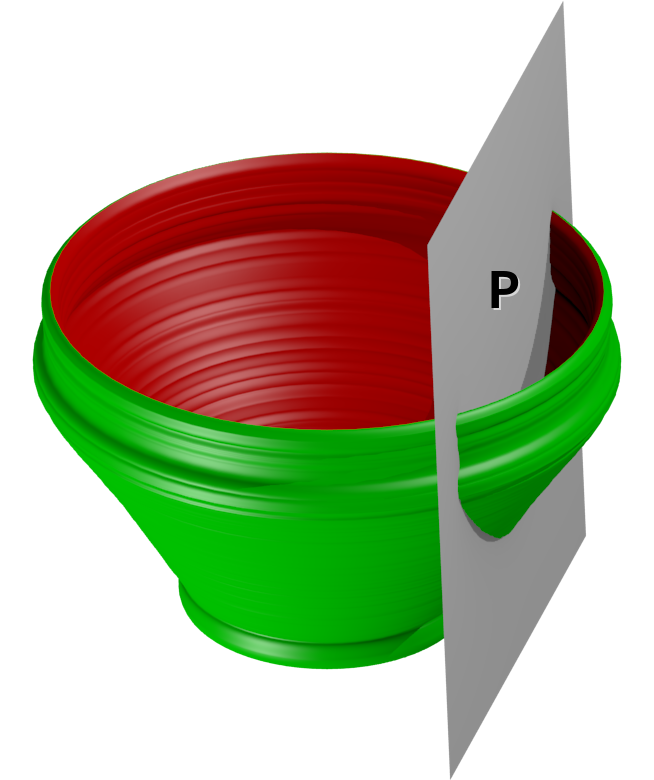}&
\includegraphics[width=0.212402431213\linewidth]{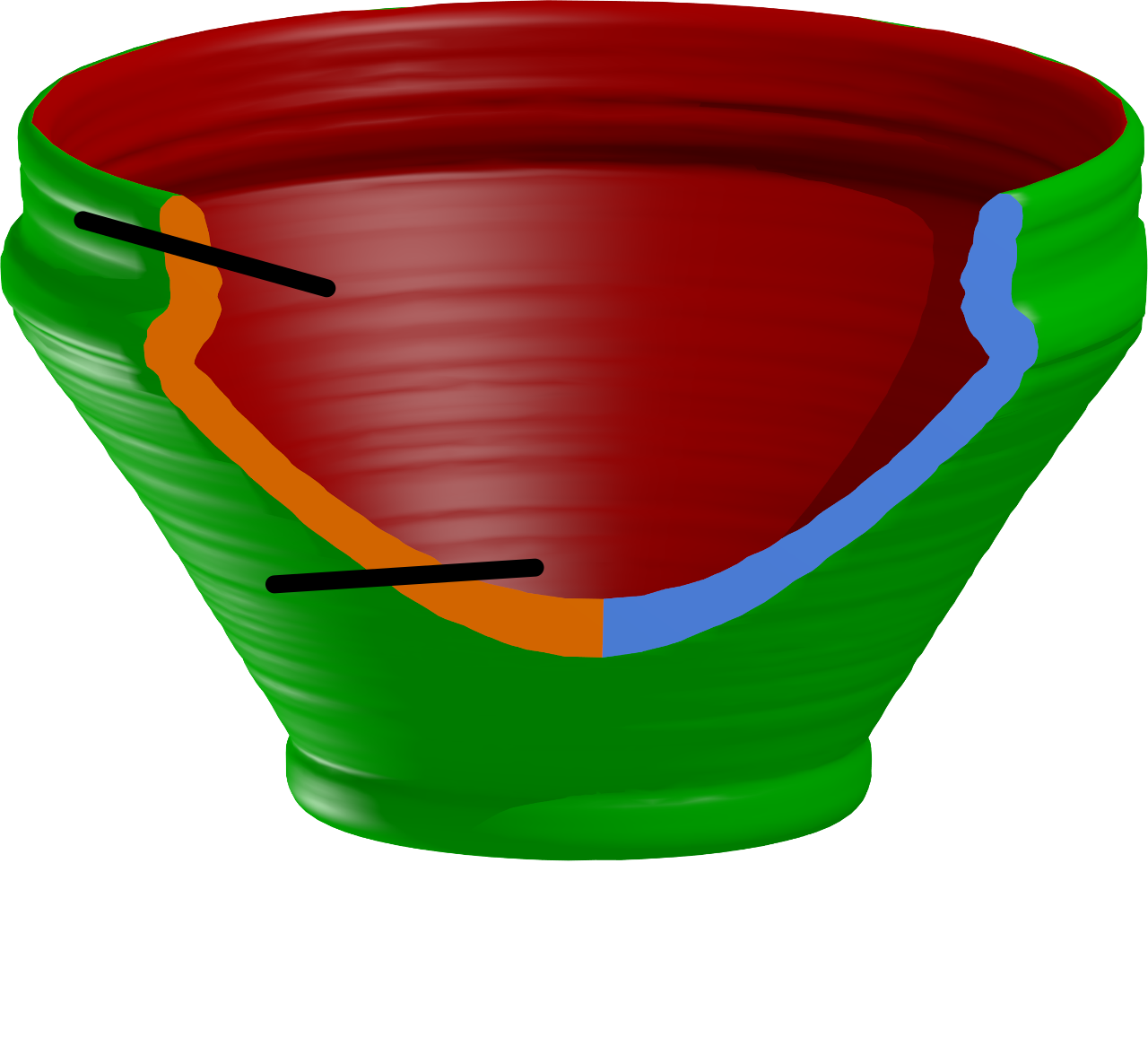}\\
(a) & (b) & (c) & (d) \\
\end{tabular}\caption{Sketch processing. (a) The processed sketch with the inner profile, outer profile, and rotation axis. (b) The rotation process. {\color{black} The inner and outer profiles are positioned for rotation around the rotation axis.} (c) A cutting plane $P$ through the 3D pottery. (d) The complete fracture face. In practice, only one of two sides (marked in orange and blue) is present in most excavated sherds. We further cut the top and bottom of the fracture, using two lines, to create a sherd with more realistic edges and size}
\label{fig:rotation}
\end{figure*}

\paragraph{Extraction of sketch lines from  catalogs.}
Extraction of the profile from the sketch is done by tracing the edges of the black area denoting the profile of the vessel (the left half of Figure~\ref{fig:leftsherdrightsketchs}(c)). Note that some profile drawings are incomplete, as they are used to capture  distinctions between subtypes that share similar vessel bodies, or because not enough is known about the full shape. For these, manual annotations mark the edges that are in fact ``missing'' (see Figure~\ref{fig:partial_profile}). Handles, if present in the profile, are removed.  Finally, the scale is extracted from the ruler present somewhere on the page (not always in the same crop as the drawings). The result of this extraction process can be seen in Figure~\ref{fig:rotation}(a).

\paragraph{Efficient generation of synthetic fracture faces.}
We propose a direct method for generating synthetic sherd outlines, without the need to reconstruct a 3D model. Our method is based on the observation that every point rotated around the rotation axis forms a circle in 3D, perpendicular to the rotation axis. We consider the sketch as placed on the $xz$ plane, with the rotation axis aligned with the $z$ axis. Every outline point $(p_x, p_y)$ defines the 3D geometric location satisfying the two equations $x^2 + y^2 = p_x^2$ and $z = p_y$; see Figure~\ref{fig:rotation}(b).

To generate a fracture, we consider a random 3D plane $P$ that intersects the model (Figure~\ref{fig:rotation}(c)), where the angle between the plane and the $z$-axis is kept small {(below $20\degree$)} to simulate the more distinctive real-world fractures, which are almost vertical. 

We then compute the intersection of the plane $P$ with the $O(n)$ circles defined above, {$n$ being the number of outline points}, an operation that can be performed in constant time per circle, thus generating an outline of a 3D sherd in linear time. Note the following: (i) The process is done for both the inner profile and the outer profile, resulting in two curves, with given annotation as either being the inside or outside. (ii) If the plane $P$ is not tangent to the circle, there are two intersection points per circle. Since most sherds are not presented with both sides of the cut (Figure~\ref{fig:rotation}(d)), we pick the same side for all intersections, by using the same root of the quadratic formula (i.e.\@ using only the positive square root of the discriminant).

To project the sherd back into 2D, we compute the plane coordinates of all the circle intersection points (relative to some arbitrary origin). To form the closed polygon, we connect the intersection points in the same order of their originating points on the profile, while skipping circles that have no intersection with the plane.
To add further realism to the generated fracture, we need to reduce its extent to match the dimensions of real potsherds. Therefore, we cut the resulting polygon using two almost-horizontal lines---one at the upper part of the polygon, and one at the lower part (also shown in Figure~\ref{fig:rotation}(d)).

The entire generation process can be done in linear time, without the need to simplify the outline, thus providing a significant improvement over the previously proposed method. The process is also much easier to implement, and does not require  2D envelope computations, fracture type analysis, and other complexities of~\cite{icdar}. Without much optimization, our implementation is fast enough to generate data on the fly, with no slowdown in the training process.

\paragraph{Outline normalization.}
All synthetic outlines are translated so as to have their center of mass at the origin. The typology of pottery is such that a certain feature may be associated with one class or another based on the size of the vessel, making the scale information crucial for proper classification. Therefore, unlike most of the recently reported applications of point clouds in the literature, we do not normalize each input to the unit sphere.

\begin{figure}[tb]
\begin{tabular}{cccc}
%\centering
\hspace{-.3cm}
\includegraphics[height=3.04cm]{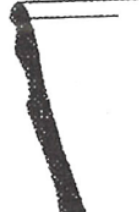}&
\includegraphics[height=3.04cm]{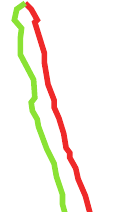}&
\includegraphics[height=3.04cm]{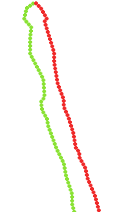}& \includegraphics[height=3.04cm]{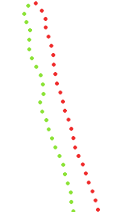}
\\
(a) & (b) & (c) & (d) \\
\end{tabular}
\caption{Propagation of artifacts as a function of sampling resolution. (a) A scan of a drawing from the catalog, depicting only the rim of a vessel, and scanned at high resolution. Printing artifacts are clearly visible. (b) Accurate tracing of the drawing propagates some of the printing artifacts as rough edges. (c) Fixed-count sampling, matching the number of points required to achieve 2mm resolution on some of the larger potsherds. Due to the sample density, the tracing artifacts are still present. (d) A resolution-limited sampling, sampling every 2mm at the scale of the real pottery. Most artifacts are no longer visible.
\label{fig:sample-resolution}}
\end{figure}

\paragraph{Point sampling.}
As we are developing a point-cloud based architecture, discrete points along the outlines from the previous steps must be sampled in order to create a suitable input. Since the drawings are scanned in high resolution, to capture as many details as possible, artifacts resulting from the printing process may be visible (see Figure~\ref{fig:sample-resolution}(a)). As a result, some of these artifacts may be reflected in the traced outline (Figure~\ref{fig:sample-resolution}(b)). Therefore,  sampling points along the outline at such fine resolutions may capture features that are mere artifacts, and not related to the actual pottery. To avoid reflecting these artifacts in the point-cloud, it is necessary to limit the sampling resolution of the outlines.

Some of the most recent point-cloud architectures, including PointNet++~\cite{qi2017pointnetplusplus} and PointCNN~\cite{hua2018pointwise}, work on a fixed number of points, while others, including PointNet, require the same number of points in each sample for training efficiently. {However,} in the case of potsherd identification, using the same number of points for all of the potsherds is detrimental, since to be able to distinguish small features we must sample along the outline at fine resolution (every 2--3 mm) on varying sizes of potsherds. Using a number of points that is sufficient to capture the details of larger sherds on smaller ones, would lead to sampling resolutions of 0.5mm (or less), which may start reflecting the printing artifacts as mentioned above, see Figure~\ref{fig:sample-resolution}(c). Furthermore, even if no such printing artifacts are present in the tracing process, learning such fine features would lead to a loss of  robustness to small defects in the clay or to tracing errors.

To overcome this issue, we allow for the sampling of fewer points in cases the outline is not long enough. We set the number of points while training to a fixed value ($K$), and always sample randomly $\min\{K, \textit{length}/\textit{resolution}\}$ points from each outline, as shown in Figure~\ref{fig:sample-resolution}(d). If fewer than $K$ points are sampled, we randomly repeat some points to reach $K$ points. The network employs max-pooling, as detailed in Section~\ref{sec:network}, and seems to be able to overcome this inconsistency in sampling. 

Finally, we note that while we restricted the training to low sampling resolutions (up to 512 points, with maximal resolution of one point every 2mm on the outline), in evaluating on real data, we use 1024 randomly sampled points, with a lower limit of 1mm on the resolution. In this way, we enjoy the efficiency benefit of training on a fixed number of points, avoiding overfitting to small details during training, while enjoying the added information at test time. 

\paragraph{Data augmentation and generalization to real data.}
When photographing potsherds, it is important that the fracture  be aligned with the image---where the sherd's vertical axis is aligned with the vertical axis of the image and the fracture surface is kept parallel to the horizontal plane, to minimize distortions in the acquired fracture shape. Note that the user has no difficulty in approximating the vertical axis $z$, since the manufacturing process creates shapes with dominant circles around $z$. The ability of the users to properly align the vertical axis (aligning both the vertical axis to the rotation axis, and the fracture surface to the image plane) was verified in field trials.

Despite the intuitive ability of archeologists to align the fracture correctly, this alignment is inexact, since it is a manual process. To achieve robustness in the face of errors in this alignment process, we simulate a small random 3D rotation (the angle is sampled from a normal distribution with $\mu=0\degree$ and $\sigma=10\degree$) on each fracture before projecting it onto a 2D outline.

Another concern with regards to  data acquisition quality arises from the nature of the field work. With one hand operating the camera and another hand holding the potsherd, a ruler that is used for inferring scale information is often left on the table and not held at the same distance as the fracture surface; see Figure~\ref{fig:leftsherdrightsketchs}(a). This seemingly small difference in distance from the camera, when combined with close range photography, has been empirically shown to lead to scale computations that cause sherds to appear up to 50\% larger than their actual size. To achieve robustness to this sort of issue, we add a random scale factor, which is sampled from a Gaussian distribution with a variance of 0.8 and a mean of 1.2 (not 1.0 since the scale extraction process is biased).

\subsection{Network Architecture}

\label{sec:network}

\begin{figure*}[tb]
\centering
\includegraphics[width=1\linewidth]{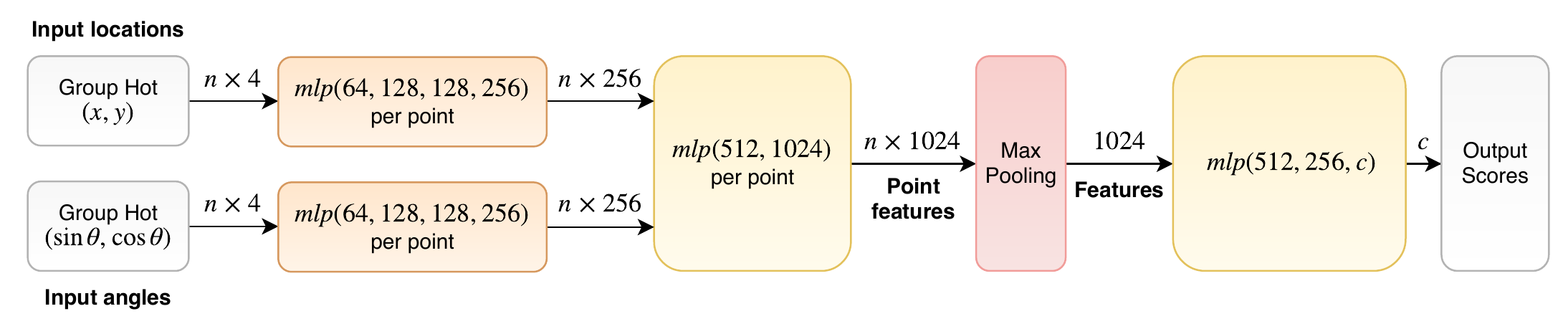}
\caption{The network architecture, which consists of two pathlines: location and angle (mlp = multilayer perceptron).}
\label{fig:arch}
\end{figure*}

Our proposed OutlineNet is based on PointNet with multiple improvements. Unlike PointCNN and PointNet++, we do not attempt to cluster points together dynamically, but rather use the natural ordering of points along the outline for enriching the available information at each point with more than just its spatial location.

In PointNet, the vector of each point goes through a series of 1D convolutions to generate a per-point feature. The network then applies a max-pooling layer to obtain a fixed size feature vector, in a manner agnostic to the order of the input points. 
In our network, we add two important pieces of information to each point: (1) annotation regarding side (inside/outside), and (2) the angle of the outline at that point, {\color{black}which gives a rotation-invariant representation of the context around the point}. Instead of representing this information as a 4-coordinate vector $(x, y, \textit{side}, \textit{angle})$, which, as we found empirically, is ineffective, we suggest a  novel approach, changing both the architecture and the data representation throughout the network. 

\subsection{Group-Hot Encoding for Side Information} The side information is a categorical value, and as such, it would typically be encoded using a one-hot encoding. However, using one-hot encoding with inputs taking continuous values can cause problems when it is significantly smaller or larger than the rest of the values. While the network can theoretically learn the proper weights to compensate for any scale, in practice this does not always work.

Instead, we suggest the following approach for combining categorical and continuous values---an approach we call ``group-hot'' encoding. To represent $d$ continuous values coupled with one categorical value with $c$ options, we suggest creating a vector $v \in \mathbb{R}^{dc}$, representing $c$ groups of $d$ values. To represent group $i$, we would zero out the values of all but the $i${th} group and store the $d$ values in that group. For our two-value categorical information (inside/outside), $(x,y)$ location values would be represented as $(x_{\textit{in}}, y_{\textit{in}}, x_{\textit{out}}, y_{\textit{out}})$ where only one pair of values is nonzero at each time.

\subsection{Multi-feature and Angle Information}

To encode the spatial context for each point, previous works construct hierarchies between  points~\cite{qi2017pointnetplusplus,hua2018pointwise}. In our case, the points are ordered, and we instead encode the immediate context around each point using angular information by considering, for every point, the cosine and sine of the angle formed at this point along the outline (encoding angle information directly with numbers suffers from the discontinuity between zero and $2\pi$).

Angular information is secondary to the spatial information, and employing representations such as $(x_{\textit{in}}$, $y_{\textit{in}}$, $\sin \theta_{\textit{in}}$, $\cos \theta_{\textit{in}}$, $x_{\textit{out}}$, $y_{\textit{out}}$, $\sin \theta_{\textit{out}}$, $\cos \theta_{\textit{out}})$ showed little to no benefit to  network performance. Instead, we employ a multi-pathway architecture to enable learning separate features for spatial- and angular-information. 

This architecture is shown in Figure~\ref{fig:arch}. We begin with two separate branches of multilayer perceptrons (MLPs): one for the angle data and one for the location data. Both branches have the same shape---the first layer has 64 hidden units, followed by two layers of 128 units, and a final layer of 256 units. The outputs of these two branches are then concatenated (per point) and fed into two perceptron layers of 512 and 1024 hidden units respectively, to obtain a feature vector of length 1024 per point. Max pooling is then performed over all points to obtain a global feature vector of the same size. Going through an additional MLP (512, 256 and $c$ hidden units) and a final softmax layer, we obtain the output scores for the $c$ classes. 
All MLPs, except for the one producing the output score, employ ReLU activations. The MLP after the max pooling employs a dropout with a rate of 0.7 after each layer, except for the last one. A batch size of 128 and the Adam optimizer~\cite{KingmaB14} with an initial learning rate of $1\times 10^{-6}$ are used for training.

\section{Decoration-Based Identification}

The drawings and the colors used to decorate pottery, can be classified based on the usage of specific colors or their combination, by the type of patterns that are being painted, by the areas that are being painted, and more. 
For appearance based classification, our work was mainly carried out on the Majolica of Montelupo pottery. The collection of the dataset was led by the University of Pisa (UNIPI), using both existing images (from archaeological excavations, PhD theses, and more) and multiple photography campaigns. The majority of the images were collected during the Autumn of 2017, with more than 8000 sherds being photographed, covering 67 genres with more than 20 sherds, many of which with more than 100 sherds. All the pictures have been classified by UNIPI.

Similar to other applications of computer vision in domains in which the data is relatively scarce, we rely on feature extraction from an existing neural network to the task at hand. As the base for our network, we use a pre-trained version of the ResNet-101 network~\cite{resnet} trained on the ImageNet collection~\cite{imagenet}. The network operates on RGB (color) images, after these have been resized to a fixed size of $224\times224$ pixels.

\begin{figure*}
    \centering
    \includegraphics[width=\textwidth]{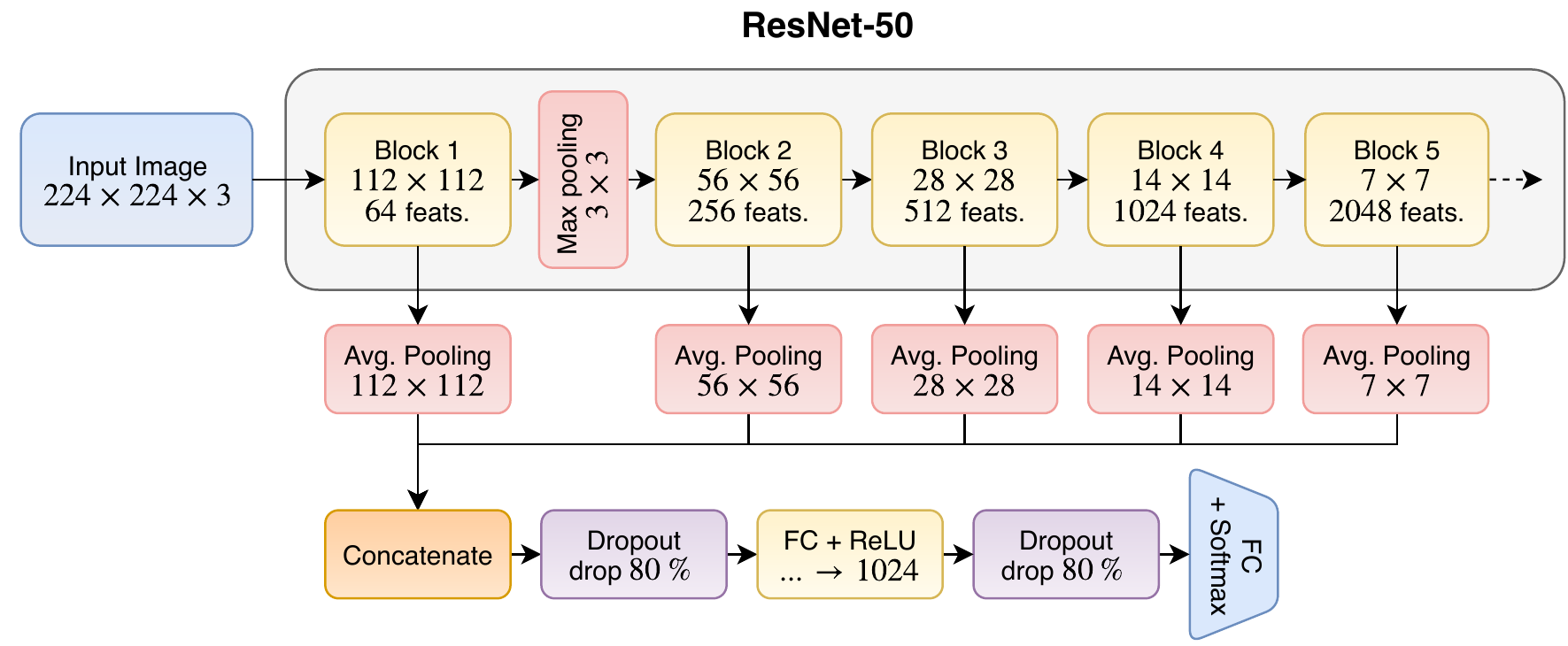}
    \caption{The updated ResNet-based network for classifying potsherds by their appearance. The ResNet part of the network is frozen and only the parts operating on the feature vectors are being trained. We use a significant dropout to reduce the overfitting that may occur with large feature vectors.}
    \label{fig:appearance/network-v2}
\end{figure*}

In order to utilize features at various levels of abstraction, we combine features from multiple levels of depth, while the lower levels would encode color and texture, the top layers encode complex patterns that are more related to the semantic content of the image. ResNet-50 is composed from a sequence of blocks and we concatenating the features from blocks 2-5 in order to obtain one large feature vector, as can be seen in Fig.~\ref{fig:appearance/network-v2}. The feature maps from each block is a multidimensional map with a varying number of channels ranging from 256 (block 2) to 2048 (block 5). Since we want to be position invariant, prior to this concatenation, we eliminate the spatial information by performing average pooling over the entire spatial extent of each channel, resulting in one vector of features from each block. To account for the different statistics of the features, we normalize each feature (in the concatenated vector) separately to have a mean of zero and a variance of 1 on the training set.

The concatenated vector contains 3840 features. To this vector, we apply a dropout regularization, a fully connected layer projecting to 1024 features, followed by a ReLU activation, a second dropout, and a projection to the number of classes followed by a softmax operator. Both dropout layers employ a high level of drop (80\%) in order to increase robustness and decrease reliance on specific features. During training, we fix the parameters of the ResNet layers that extract the features, and only train the parameters of the fully connected layers on top of the features. 

To fit the images to the expected input dimensions of the ResNet model, we scale them to 224 pixels (along the shorter axis), and crop them (equally on each side of the longer axis), to obtain a $224x224$ image. To train our network to work with varying amounts of decorations/background inside the image, we enrich the original image dataset by adding augmented versions of each image: for each image, we scale it to four different sizes; on each scaled image, we create three flipped versions (unflipped, horizontally flipped and vertically flipped); we crop all of those images, leaving just the center square. Thus from each image, we create 12 images that can go into the neural network, increasing the dataset size from around 8000 images to about 100,000 images. 

In our initial experiments, varying illumination was the most challenging factor in identification. To solve this lack of robustness, we simulating different white balance results, and various brightness and contrast adjustments. This was applied during the generation of the training dataset by multiplying the luminosity (``brightness'') of all the pixels within each image, using a randomized factor (between 0.8 and 1.2) per image, to simulate different lighting conditions. To compensate for different white balance setups, we additionally apply a similar random multiplicative factor to each channel in the image; that is, we multiply each of the Red/Green/Blue channels, by a separate random constant factor, to change the ratio between colors in the image.

\begin{figure}
  \begin{center}
    \includegraphics[width=0.2\textwidth]{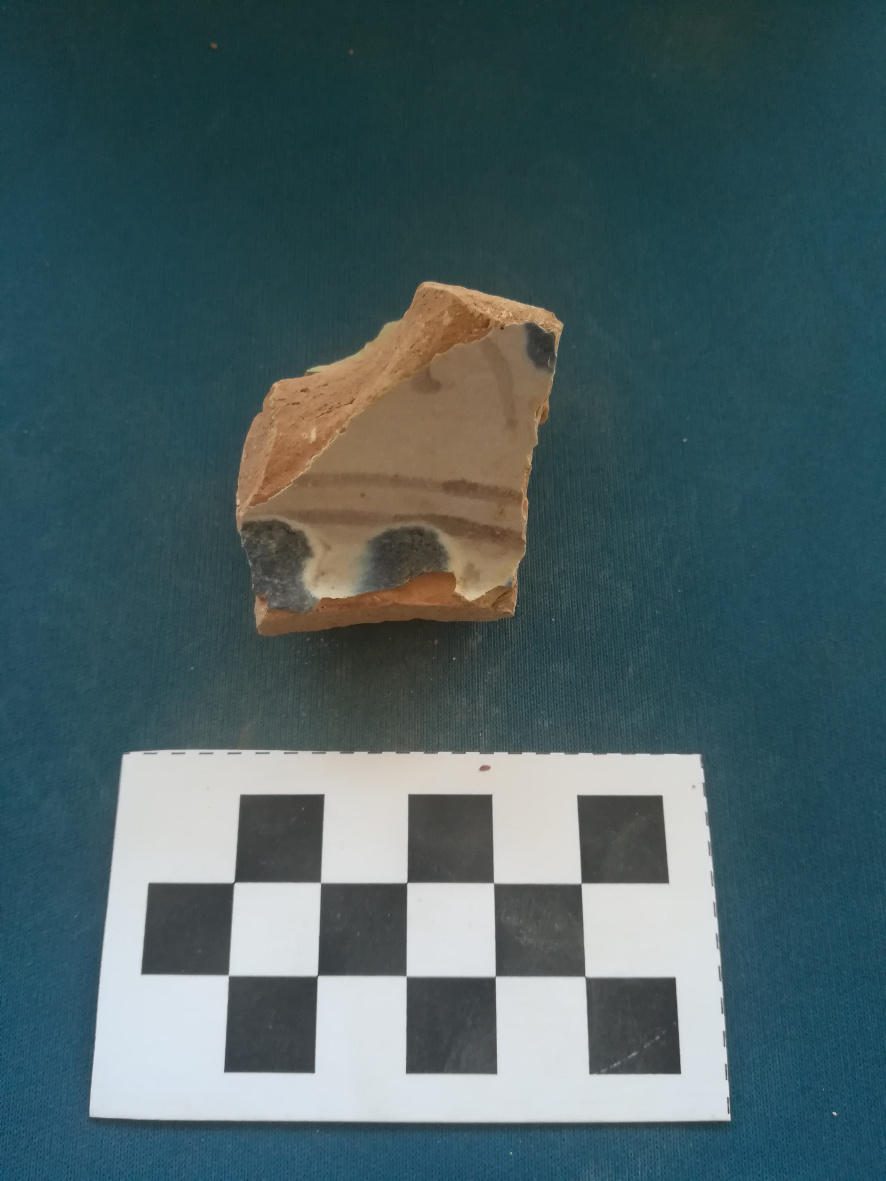}
    \includegraphics[width=0.2\textwidth]{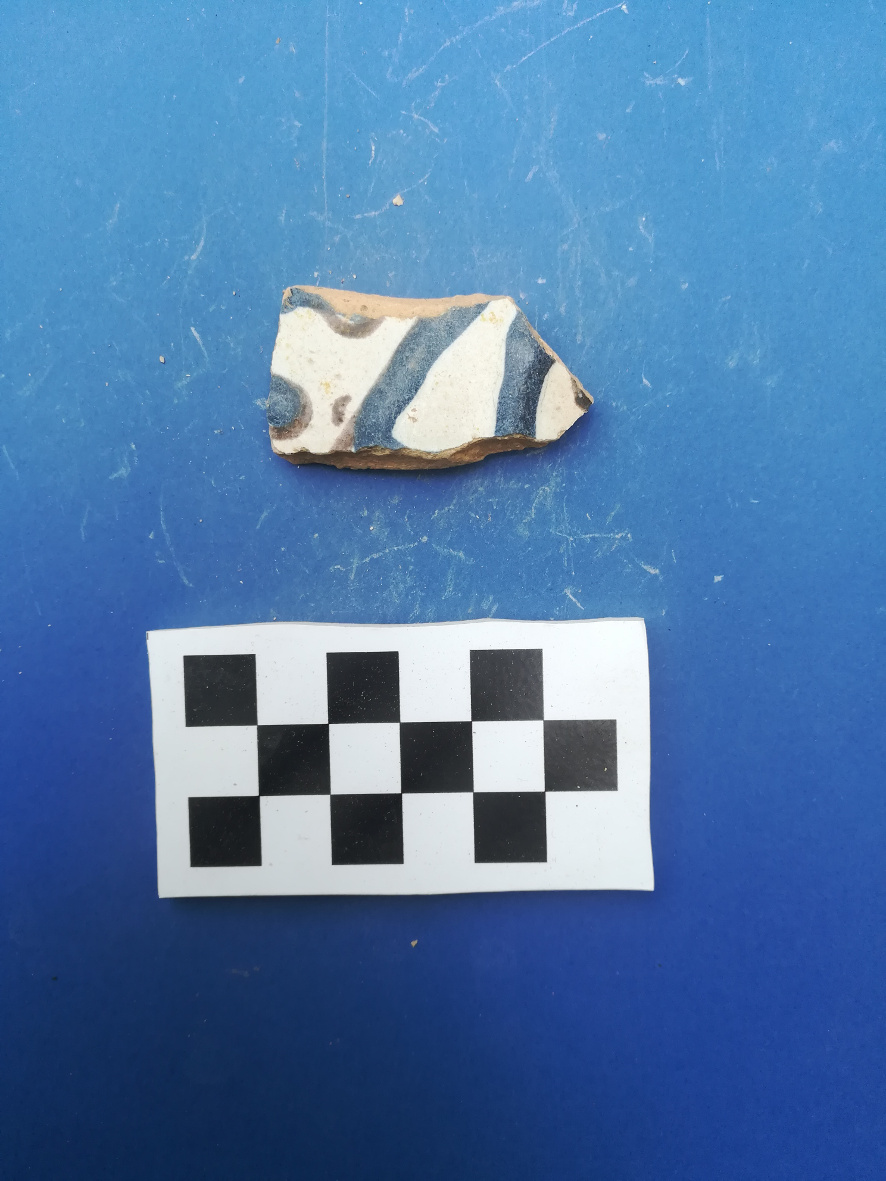}
    \includegraphics[width=0.2\textwidth]{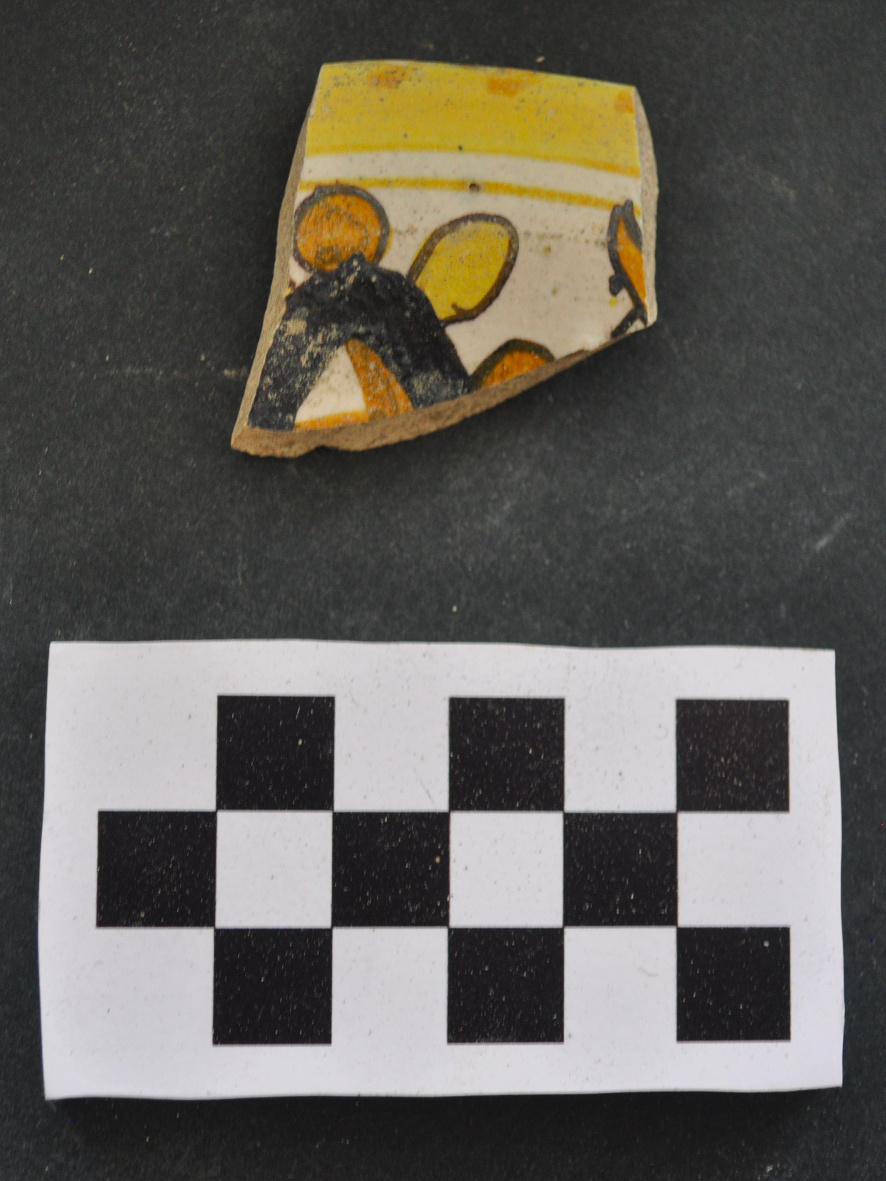}
  \end{center}
  \caption{Three typical images captured during the photography campaign. }
  \label{fig:appearance/sherds-with-bg}
\end{figure}

In addition, the imaging conditioned (background, ruler) varied considerably between the collection campaigns and the other sources, leading to an inherent bias, 
as can be seen in Fig.~\ref{fig:appearance/sherds-with-bg}. To overcome this, we extract the foreground of the training images automatically. During test, the GrabCut algorithm~\cite{Rother:2004:GIF:1186562.1015720} is used to extract the relevant image part.

\begin{figure}
  \begin{center}
    \includegraphics[width=0.2\textwidth]{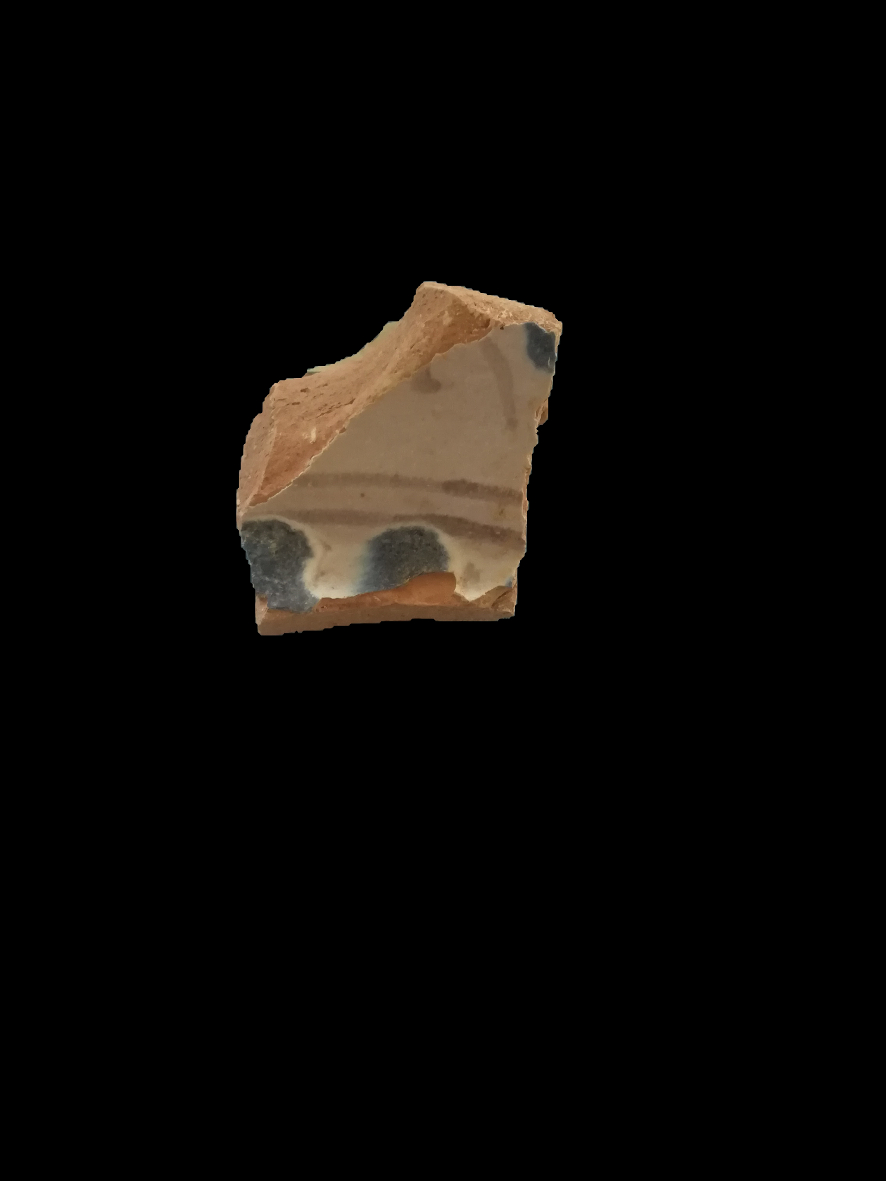}
    \includegraphics[width=0.2\textwidth]{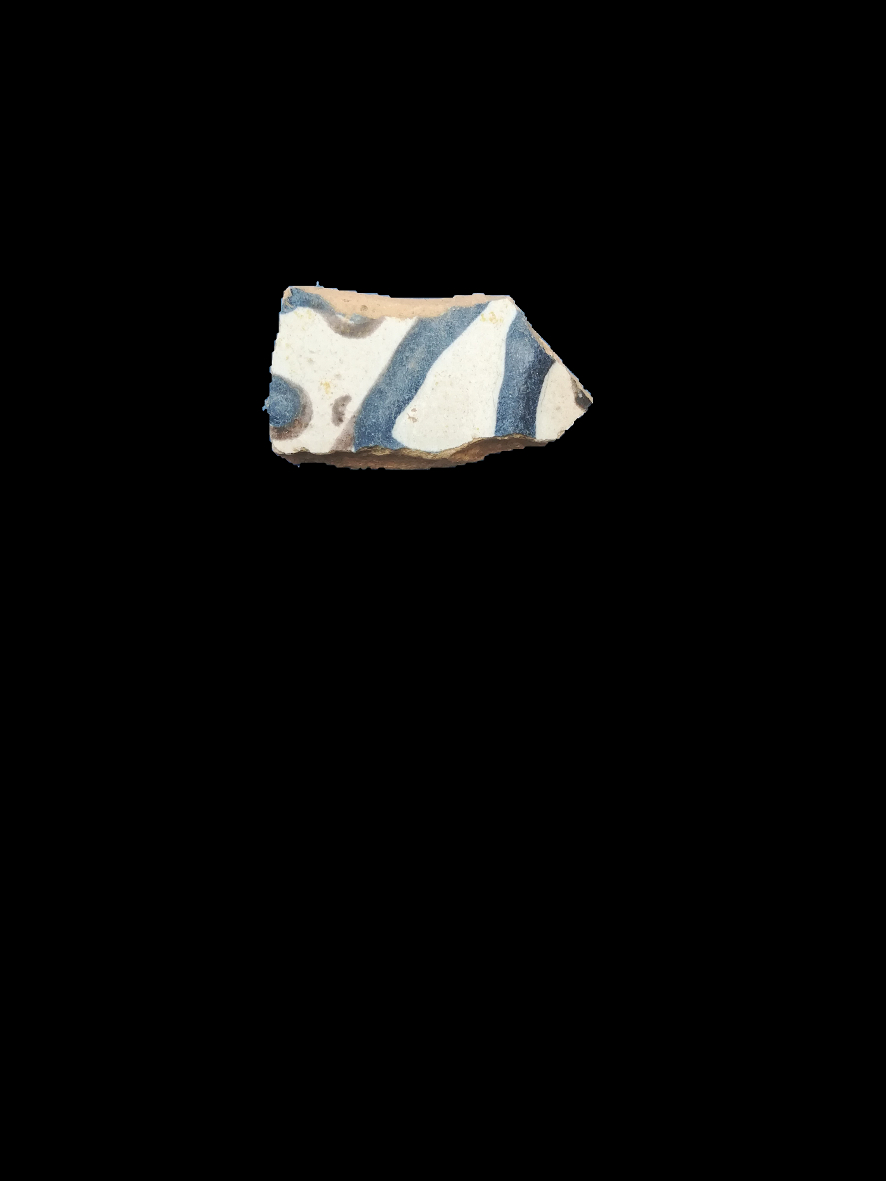}
    \includegraphics[width=0.2\textwidth]{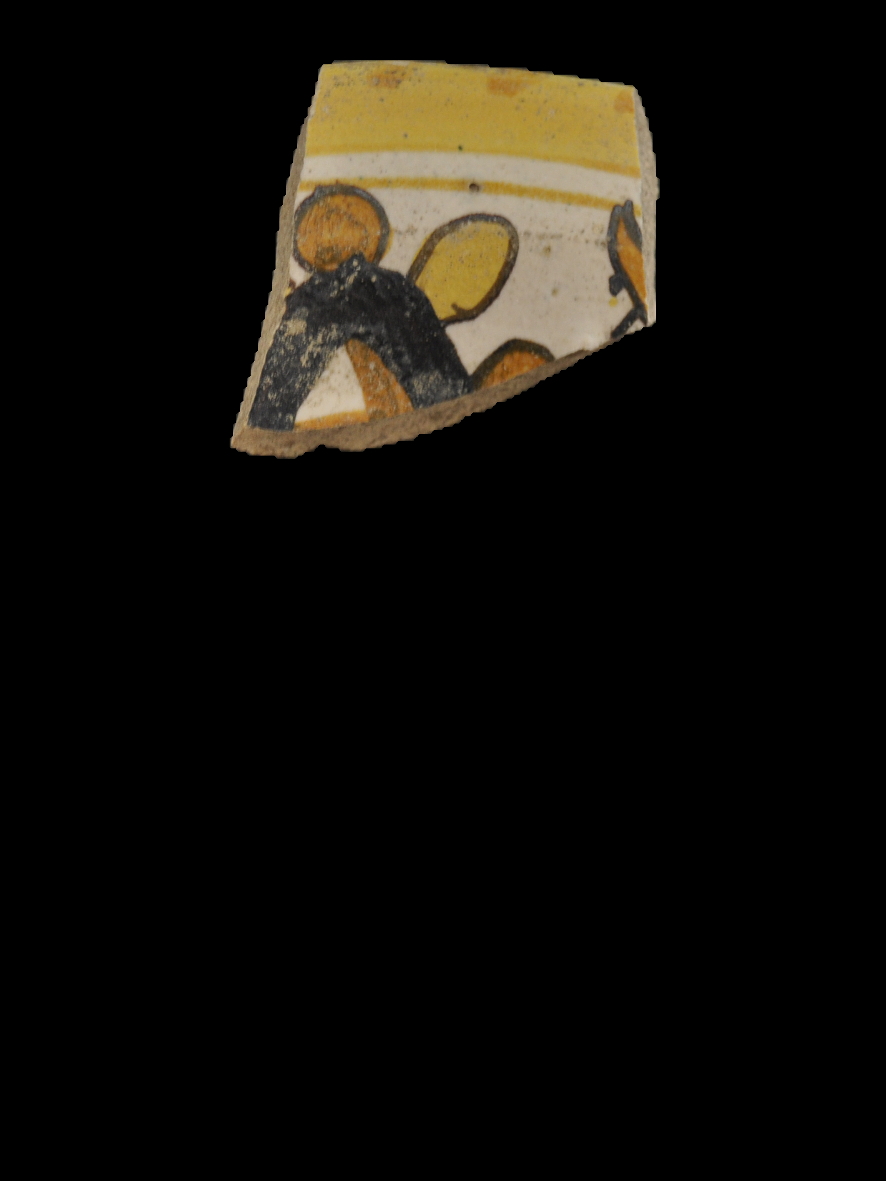}
  \end{center}
  \caption{Three typical images captured during the photography campaign}
  \label{fig:appearance/sherds-without-bg}
\end{figure}

\begin{figure}
  \begin{center}
    \includegraphics[width=0.2\textwidth]{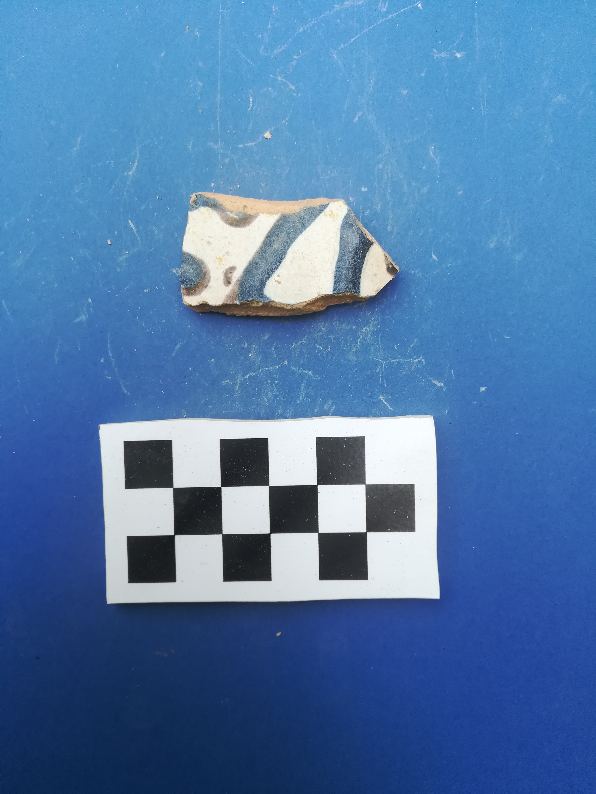}
    \includegraphics[width=0.2\textwidth]{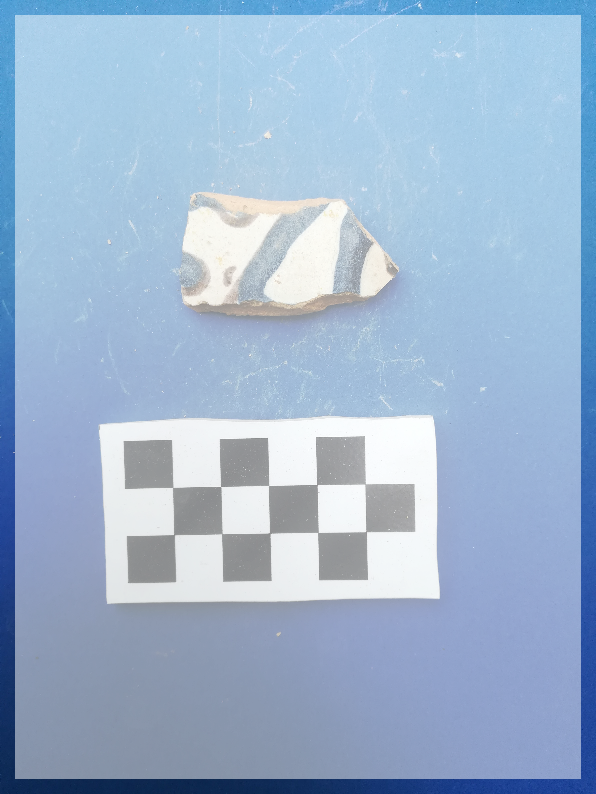}
    \includegraphics[width=0.2\textwidth]{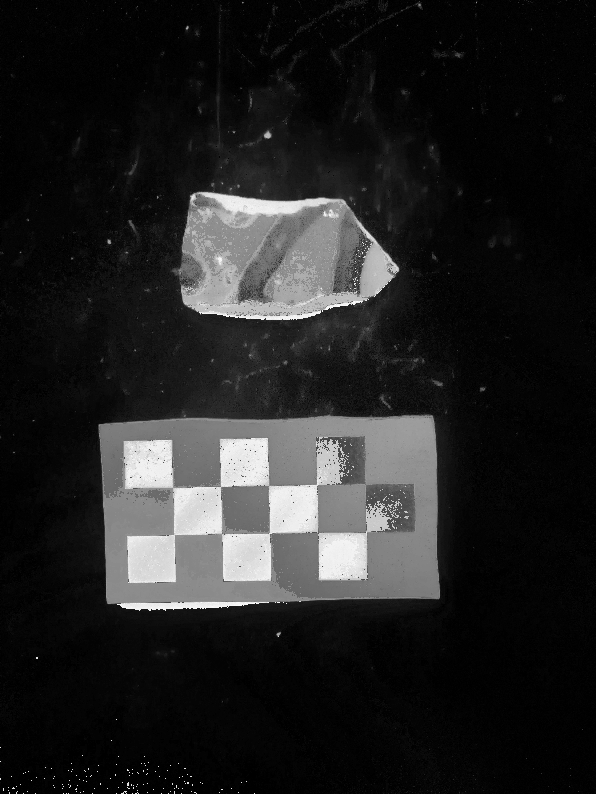}
    \includegraphics[width=0.2\textwidth]{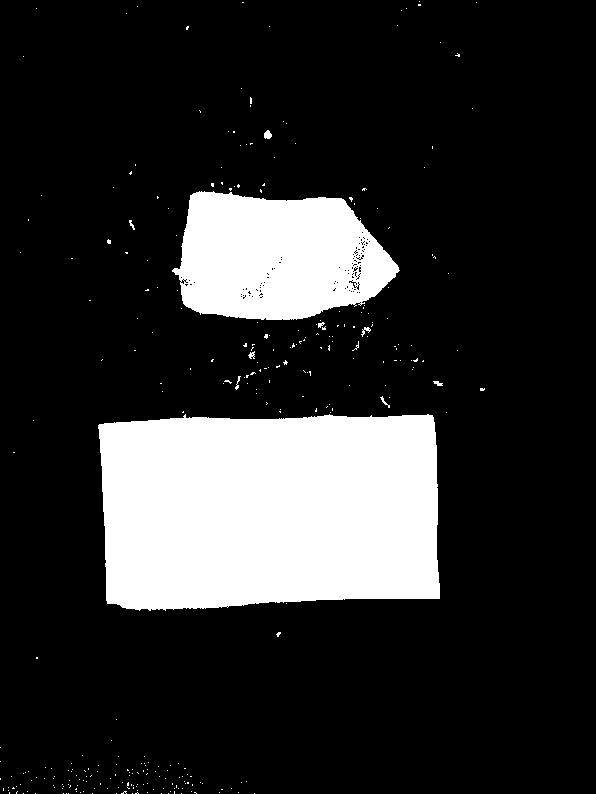}
  \end{center}
  \caption{Visualizing the first two steps of the manual foreground extraction visualization: First we sample colors from the edges of the image, and then we compute the minimal distance between each pixel to any edge pixel. Finally, we threshold the image until we get two large islands, that can be further cleaned with morphological operations}
  \label{fig:appearance/fg-extraction}
\end{figure}

Most images in the dataset are similar to those in Fig.~\ref{fig:appearance/sherds-with-bg} - the images have a mostly uniform background, a potsherd and a ruler. For these images, we devised a heuristic that can provide an automatic input for the GrabCut algorithm, instead of requiring manual input: (i) We obtain color samples from the background, we sample pixels around the border of the image, relying on the fact that both the ruler and the potsherd are centered in the image, and not touching the image edges. (ii) Using the samples of background color, we measure the distance of each pixel in the image from the ``nearest'' background color we sampled. (iii) On this distance image, we apply a threshold operation (everything below the threshold is black, everything above is white). The minimal threshold is used such that there are exactly two white connected components, aimed to be the ruler and the potsherd (see Fig.~\ref{fig:appearance/fg-extraction}). (iv) To make sure the image was indeed segmented correctly, we check whether one of these islands is a ruler. Specifically, we use a heuristic that is based on applying the Harris corner detector~\cite{harris} to identify the corners of the checkerboard pattern. (v) On the segmented area containing the potsherd, we apply GrabCut to obtain a finer segmentation, initialized according to the segment's boundaries.

\section{Loss Reweighting}

Most common techniques for combating low class accuracy introduce weights on the loss expressions of individual samples, with higher weights assigned to inputs from classes with low accuracy. While the rationale in these techniques is clear, there is no actual guarantee that it will make the classifier learn anything ``meaningful'' about these classes. For example, one way to push the accuracy of a given class upwards is to increase the bias of the logit to this class. While this uniformly increases the chance of all inputs to be classified into the class (including correct inputs, thus increasing the accuracy), this new classifier does not contain any new information compared to the previous classifier. This phenomenon, which can be identified by an accuracy increase that is accompanied by an increase the number of false-positives that are predicted to be in this class, was encountered multiple times during our research.

To mitigate the issue described above, we propose a new loss function, one that weights samples not just by their true label but also by their predicted label. For each sample, the loss has one weight by the true label (assigning higher weights for classes with low accuracy) and another weight by the predicted label, assigning higher weights to mis-classifications. The second weight is aimed at tackling an increase in accuracy, which is accompanied by an increase in the number of false positives. 

As it turns out, the new loss proposed not only increases the uniformity of the accuracy among the classes, but also increases the overall performance on the test set. We attribute this to the fact that during testing, the same types of confusions that occur in the training data are likely to occur, only more frequently.

Let $X = (x_1, x_2, \ldots , x_n)$ be the set of inputs to a specific batch, and let $Y = (y_1, y_2, \ldots , y_n)$ be their respective labels, where $y_i \in \{1, 2, \ldots , c\}$. Let $f(x_i)=(f^1(x_i), \ldots , f^c(x_i))$ be the probability vector predicted by the model $f$, and let $\widehat{y}_i = \argmax{f(x_i)}$ be the class predicted by the network for input sample $x_i$.
As the underlying loss, we employ the conventional cross entropy loss $\ell_i(f)=-\log f^{y_i}(x_i)$ for classifier $f$ and input sample $i$. The new per-sample loss function, called \emph{CareLoss} as an antonym of neglecting some classes, is denoted  $\tilde{\ell}_i(f)$. It is created by weighting $\ell_i(f)$ by two weights $u$ and $v$, which are associated with its ground truth label and the predicted label, respectively:
\begin{align*}
\tilde{\ell_i}(f) := \tilde{\ell} (f, y_i, \widehat{y}_i) := u(f, y_i) v(f, y_i,\widehat{y}_i) \ell_i (f)
\end{align*}
where
\begin{align*}
\widehat{u}(f, y_i) &:= \exp(- \alpha_u \psi(f,y_i))\\  
\widehat{v}(f, \widehat{y}_i) &:= \exp(+ \alpha_v \rho(f,\widehat{y}_i))\\ 
u(f, y_i) &:= \frac{\widehat{u}(f, y_i)}{\sum_{j}\widehat{u}(f, j)} \\ 
v(f, y_i, \widehat{y}_i) &:=  \frac{1}{\eta}\left ( 1 + I_\text{miss}(y_i, \widehat{y}_i)\frac{\widehat{v}(f, \widehat{y}_i)}{\sum_{j}\widehat{v}(f, j)}\right )
\\
I_\text{miss}(y_i, \widehat{y}_i) &:= \begin{cases}
1 &y_i \neq \widehat{y}_i \\
0 &\text{otherwise}
\end{cases}
\end{align*}
Here, $\alpha_u$ and $\alpha_v$ are positive parameters, $\psi(f,j)$ is the accuracy of the classifier $f$ over the inputs originating from class $j$, $\rho(f,j)$ is the false-positive rate of the classifier $f$ into class $j$ (the ratio between the number of samples that are classified falsely into class $j$ and the total number of misclassified samples), and $\eta$ is a normalization parameter that makes sure that, per batch, all of the $v$ terms sum to one. In other words, we define unnormalized weights $\widehat u$ and $\widehat v$ based on the accuracy of the true label and the prevalence of the false positive cases that result in the predicted label and then convert these to pseudo probabilities $u$ and $v$. The indicator $I_\text{miss}$ is 1 for a sample that is classified incorrectly, 0 otherwise, and the weight $v\neq 1/\eta$ only for samples that are misclassified. 

Note that the signs are such that we weigh up samples from classes with low accuracy and samples that are falsely predicted to be of classes with high false positive rate. We especially pay attention to samples of neglected classes that are mapped to one of the classes that are often predicted.

Both $\psi(f,j)$ and $\rho(f,j)$ are computed empirically: $\psi(f,j)$ is the ratio of training samples from class $j$ that were classified as class $j$, and $\rho(f,j)$ is the ratio of misclassified training samples that were incorrectly classified as class $j$. During training, the weights $u,v$ are updated periodically every $b$ batches, using a moving average with momentum $\gamma$ to avoid sharp changes in the loss function. The underlying class accuracies are recorded over the $b$ batches, and the counters are reset after every weight update, in order to reflect an updated state of the classification confusion. The parameters we used are $b=50$, $\gamma = 0.8$, $\alpha_u=6$, and $\alpha_v=5$.

\section{Experiments}

While most reported methods in the literature employ a test data that is available during the development stages, this may cause  overly optimistic results due to multiple hypotheses testing and other biases. This poses an increased danger in domains in which the dataset are not always large, e.g., in medicine, natural  sciences, and, needless to say, archeology. 

The development of the reference tool was planned as a two phase process, which includes a lengthy validation process. In the first phase, the methods were developed on potsherds of one family, collected from the same geographical region. In the second phase, additional datasets were provided, each with its own set of classes and defined train and test splits.

\subsection{Phase I Experiments}

In the first phase, the classification task is to classify potsherds of \emph{Terra Sigillata Italica (TSI)} into one of 65 standardized top-level classes defined in the Conspectus catalog~\cite{conspectus}. These top level classes are defined by 435 sketches, and each class has between 1--8 associated sketches, from which class-balanced synthetic data is generated. 
The outlines of the real-world sherds, used exclusively for testing, were extracted from images collected across Europe using a dedicated mobile app. 

As part of the outline extraction, the user annotates the outline segments with the inner or outer side information, which is easily inferred by the archeologists.  Due to technical issues with the development of the mobile app, the annotation of the outline is currently a manual process in which the user simply taps on points along the outline and extracts a coarse polygon. This makes the dataset even more challenging---due to potential lack of fine details, and inaccuracies resulting from a touch-based input. Since accurate extraction is important for the task at hand, semi-automatic algorithms such as Intelligent Scissors~\cite{DBLP:conf/siggraph/MortensenB95} and GrabCut~\cite{grabcut14}, which strike a balance between efficient extraction  and fine-grained user control, are in the process of being added to the mobile app.

\begin{table*}
\centering
\caption{TSI dataset results. Mean classification accuracy across all classes for OutlineNet and various alternatives.}
\label{tab:cancercorr}
%\begin{small}
\begin{tabular}{ld{1}d{1}d{1}d{1}d{1}d{1}}
\hline
& \multicolumn{2}{c}{Synthetic data}& \multicolumn{4}{c}{Real-world test data}\\
\cline{2-3}
\cline{4-7}
Method	& \multicolumn{1}{c}{Train} & \multicolumn{1}{c}{Test} & \multicolumn{1}{c}{Top 1} & \multicolumn{1}{c}{Top 2} & \multicolumn{1}{c}{Top 5} & \multicolumn{1}{c}{Top 10}\\
\hline
OutlineNet & 60.9\% & 70.0\% & 22.0\% & 32.7\% & 57.9\% & 73.7\% \\
%{\color{black}ONLY $v$ @ RACK-G03} & \\
PointNet 8D features & 54.4\% & 71.3\% & 16.4\% & 26.9\% & 44.5\% & 65.2\% \\
%{\color{black}ONLY $v$ @ RACK-G05} & \\
PointNet 2D points & 50.5\% & 71.1\% & 23.1\% & 31.3\% & 52.4\% & 72.9\% \\
%{\color{black}ONLY $v$ @ RACK-G05} & \\
PointCNN 8D features & 23.8\% & 2.9\% & 0.0\% & 1.3\% & 2.2\% & 7.1\% \\
%{\color{black}ONLY $v$ @ RACK-G05} & \\
PointCNN 2D points & 45.4\% & 2.8\% & 0.0\% & 3.8\% & 9.7\% & 19.4\% \\
%{\color{black}ONLY $v$ @ RACK-G05}& \\
PointCNN 2D points, unit radius & 23.7\% & 2.8\% & 2.2\% & 2.7\% & 9.6\% & 14.8\% \\
\hline
OutlineNet w/o CareLoss & 63.6\% & 74.1\% & 21.8\% & 33.5\% & 51.5\% & 70.3\% \\
% Old run & 57.4\% & 65.7\% & 17.2\% & 27.2\% & 43.6\% & 61.9\% \\ 
PointNet w/o CareLoss, 8D features & 57.9\% & 72.9\% & 12.8\% & 22.6\% & 41.9\% & 61.3\% \\
PointNet w/o CareLoss, 2D points & 49.9\% & 68.7\% & 19.0\% & 28.4\% & 50.8\% & 71.1\% \\
%PointCNN w/o CareLoss, 8D features & 56.2\% & 2.7\% & 0.0\% & 0.0\% & 3.8\% & 10.3\% \\
%PointCNN w/o CareLoss, 2D points & 36.7\% & 2.7\% & 0.0\% & 1.3\% & 3.8\% & 3.8\% \\
\hline
OutlineNet with CareLoss reweighting only with $u$ & 57.6\% & 67.3\% & 21.6\% & 31.4\% & 50.0\% & 68.2\% \\
%OutlineNet with CareLoss reweighting only with $v$ & 62.5\% & 75.0\% & 25.2\% & 34.9\% & 54.2\% & 75.9\% \\
OutlineNet with CareLoss reweighting only with $v$ & \color{black}60.0\% & \color{black}70.8\% & \color{black}21.5\% & \color{black}31.0\% & \color{black}49.4\% & \color{black}67.5\% \\
OutlineNet with Focal Loss~\cite{lin2018focal} & 57.2\% & 69.6\% & 17.3\% & 28.3\% & 48.2\% & 61.1\% \\
OutlineNet with Focal Loss + CareLoss & 57.1\% & 67.3\% & 20.2\% & 28.1\% & 46.3\% & 66.5\% \\
%\color{black} OutlineNet with Focal Loss + Only $v$ & \color{black}57.6\% & \color{black}70.3\% & \color{black}16.0\% & \color{black}27.3\% & \color{black}45.8\% & \color{black}65.4\% \\
\hline
OutlineNet w/o separation of in/out & 52.9\% & 69.6\% & 22.8\% & 32.0\% & 50.4\% & 65.7\% \\
%\color{black} Ditto, only $v$ & \color{black}53.4\% & \color{black}71.5\% & \color{black}18.8\% & \color{black}33.7\% &\color{black} 51.0\% &\color{black} 68.1\% \\
OutlineNet w/o angle information & 46.8\% & 69.4\% & 20.3\% & 30.1\% & 49.5\% & 66.7\% \\
%\color{black} Ditto, only $v$ & \color{black}52.3\% & \color{black}73.8\% & \color{black}15.2\% & \color{black}27.0\% & \color{black}46.2\% & \color{black}66.0\%  \\
OutlineNet w/o group-hot encoding & 53.0\% & 66.3\% & 19.7\% & 28.5\% & 47.9\% & 65.0\% \\
%\color{black} Ditto, only $v$ & \color{black}55.1\% & \color{black}70.9\% & \color{black}18.2\% & \color{black}27.5\% & \color{black}44.5\% & \color{black}64.4\% \\
OutlineNet w/ a single pipeline & 49.1\% & 68.8\% & 18.6\% & 27.6\% & 48.9\% & 66.4\% \\
%\color{black} Ditto, only $v$ & \color{black}52.2\% & \color{black}69.2\% & \color{black}16.2\% & \color{black}23.8\% & \color{black}46.8\% & \color{black}66.0\% \\
%\hline
%OutlineNet w/o upsampling of test data & \\
\hline
OutlineNet w/o data augmentation & 86.0\% & 84.6\% & 22.9\% & 32.7\% & 53.1\% & 67.3\% \\
%\color{black} Ditto, only $v$ & \color{black}87.2\% & \color{black}88.5\% & \color{black}24.2\% & \color{black}36.6\% & \color{black}55.7\% & \color{black}66.6\% \\
\hline
\end{tabular}
%\end{small}
\end{table*}

\begin{table*}
\caption{Standard deviation of classification accuracy across all classes for OutlineNet and various alternatives on the TSI dataset.}
\label{tab:sd}
\centering
%\begin{small}
\begin{tabular}{ld{1}d{1}d{1}d{1}d{1}d{1}}
\hline
& \multicolumn{2}{c}{Synthetic data}& \multicolumn{4}{c}{Real-world test data}\\
\cline{2-3}
\cline{4-7}
Method	& \multicolumn{1}{c}{Train} & \multicolumn{1}{c}{Test} & \multicolumn{1}{c}{Top 1} & \multicolumn{1}{c}{Top 2} & \multicolumn{1}{c}{Top 5} & \multicolumn{1}{c}{Top  10}\\
\hline
OutlineNet & 11.8\% & 21.7\% & 25.1\% & 29.4\% & 30.9\% & 28.7\% \\
OutlineNet w/o CareLoss & 15.5\% & 23.8\% & 23.4\% & 31.8\% & 35.1\% & 31.1\% \\
OutlineNet w/o data augmentation & 7.4\% & 16.3\% & 25.9\% &  31.6\% &  34.2\% & 32.3\% \\
\hline
PointNet 8D features & 11.4\% & 19.1\% & 22.6\% & 30.7\% & 34.5\% & 33.8\% \\ 
PointNet 8D features w/o CareLoss & 16.9\% & 24.6\% & 19.9\% & 29.4\% & 34.0\% & 35.7\% \\
PointNet 2D points & 10.4\% & 20.2\% &27.7\% & 31.8\% & 30.7\% & 28.8\% \\
PointNet 2D points w/o CareLoss & 17.1\% & 25.1\% & 27.7\% & 31.2\% & 30.3\% & 29.6\% \\
\hline
OutlineNet with CareLoss reweighting only with $u$ & 11.3\% & 20.4\% & 26.8\% & 31.9\% & 32.6\% & 33.3\% \\
OutlineNet with CareLoss reweighting only with $v$ & 15.7\% & 24.8\% & 25.1\% & 30.3\% & 34.1\% & 29.9\% \\
OutlineNet with Focal Loss & 17.1\% & 25.7\% & 23.4\% & 27.9\% & 35.1\% & 34.1\% \\
OutlineNet with Focal Loss + CareLoss  & 11.5\% & 21.7\% & 23.5\% & 27.6\% & 32.3\% & 31.8\% \\
\hline
\end{tabular}
%\end{small}
\end{table*}

\begin{figure*}[t]
\centering
\resizebox{.94687425\textwidth}{!}{% <------ Don't forget this %
\begin{tabular}{c|ccccc|c}
\hline
Input image & 1st & 2nd & 3rd & 4th & 5th & Ground truth \\
\hline
\includegraphics[width=0.12\linewidth]{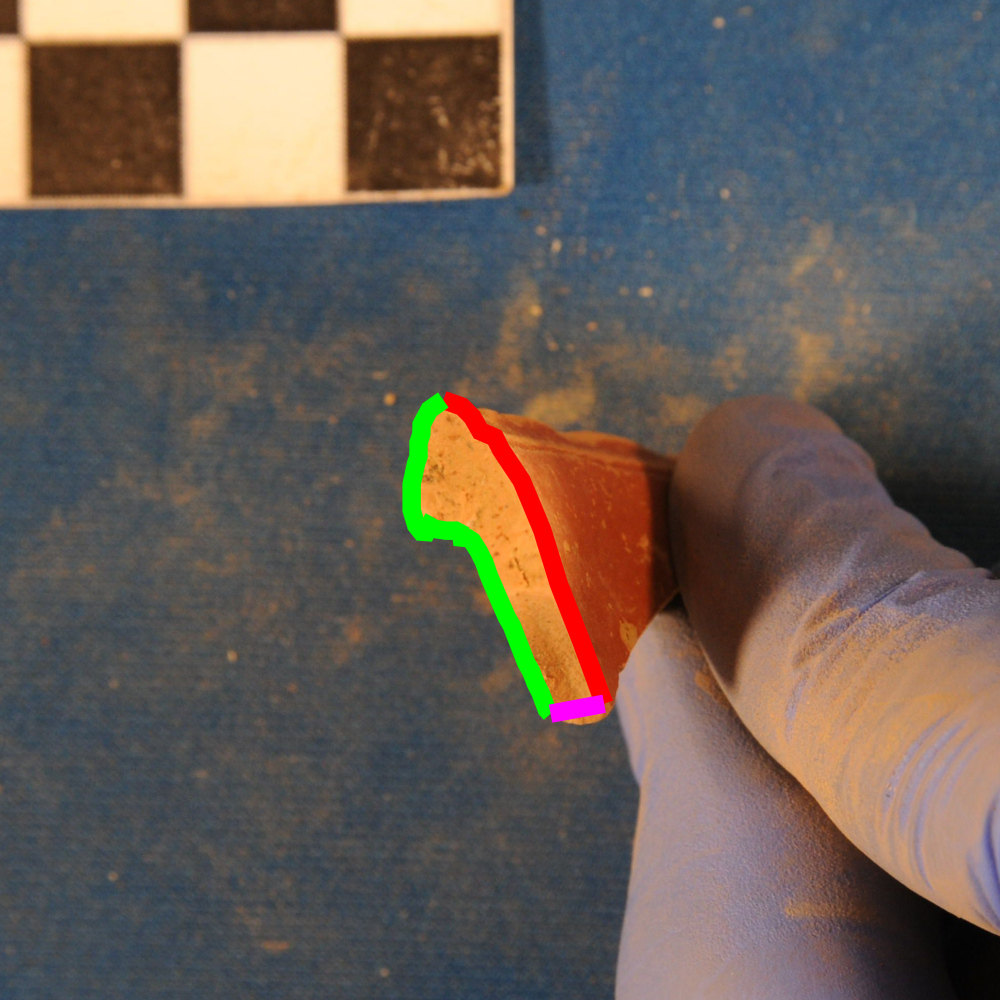} &
\hspace{1.1cm}\raisebox{1.5cm}{~~~~~~~~~~~~~~~~~~~~~~~~~~~~~~~~~~(*)\hspace{-1.40cm}}
\includegraphics[width=0.12\linewidth]{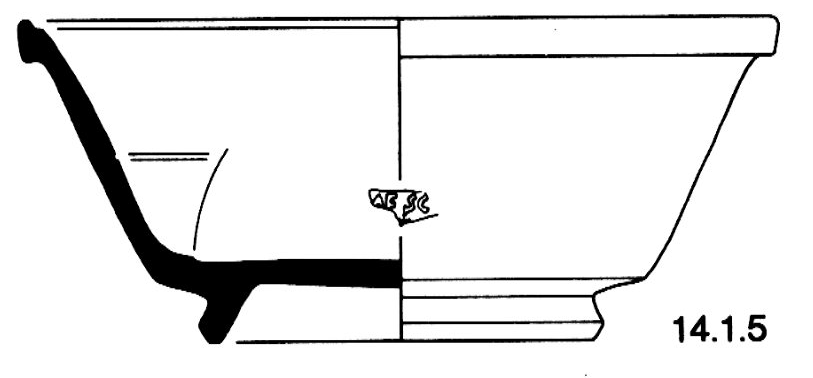} &
\includegraphics[width=0.12\linewidth]{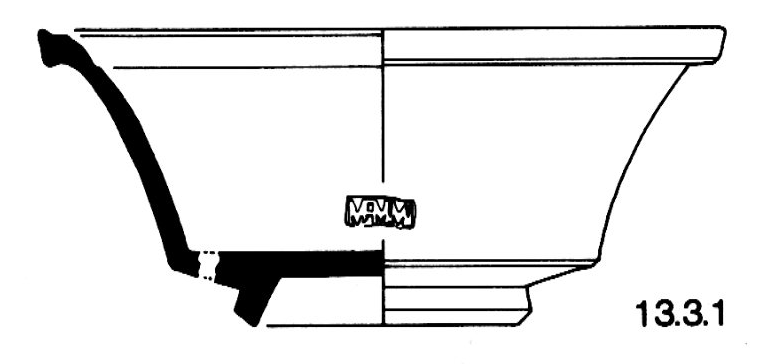} &
\includegraphics[width=0.12\linewidth]{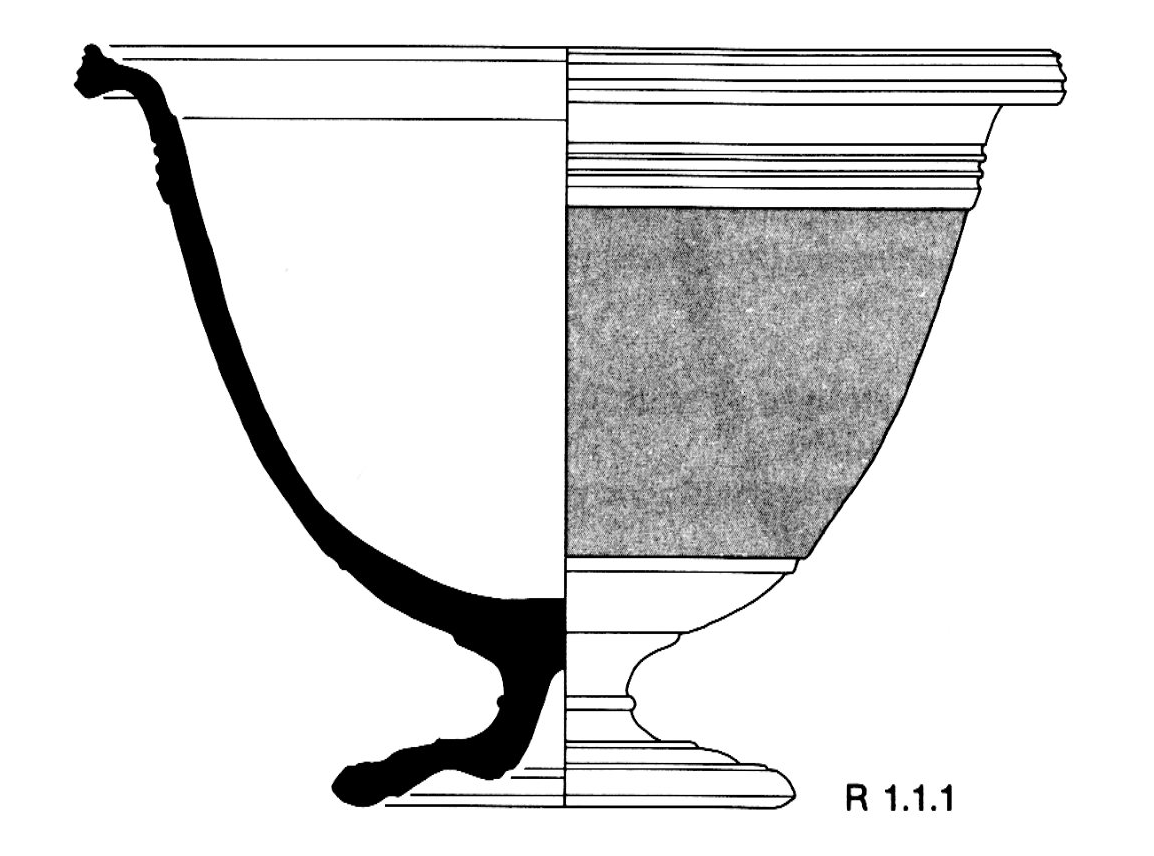} &
\includegraphics[width=0.12\linewidth]{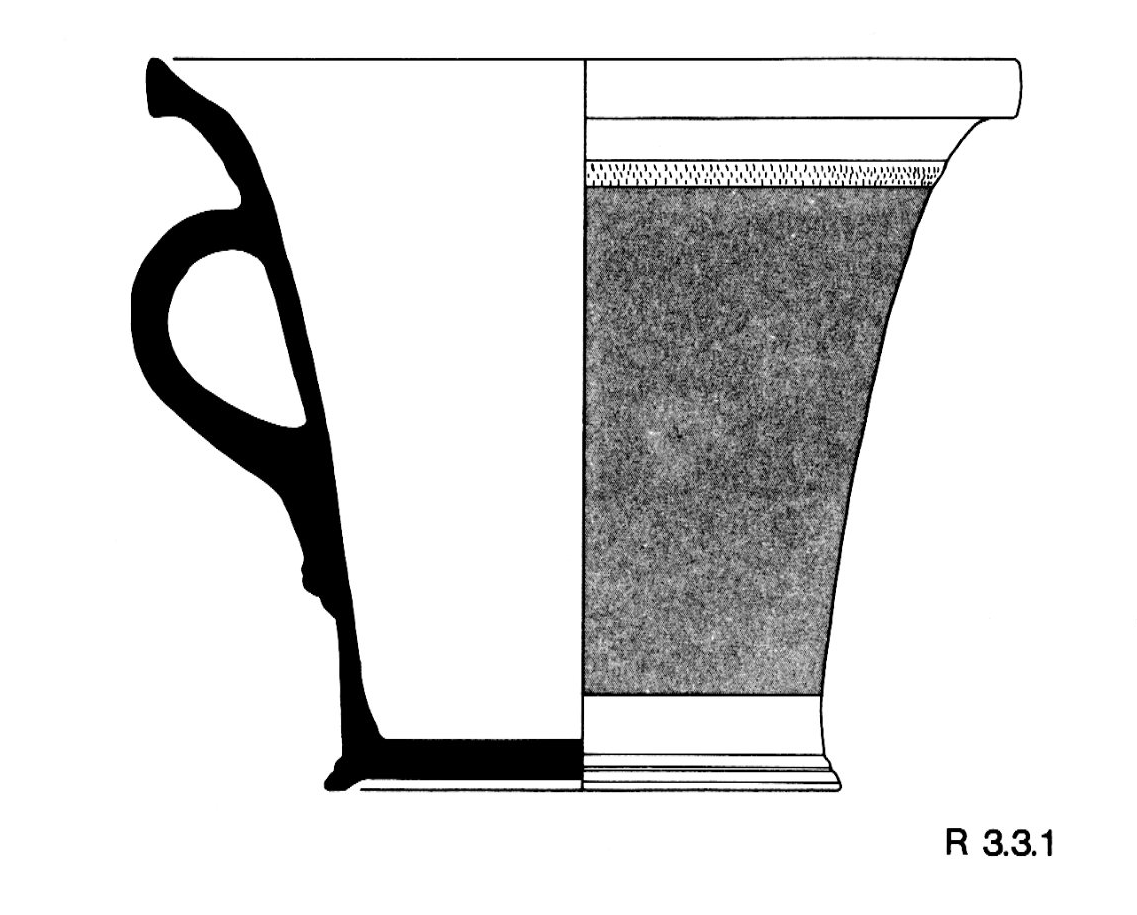} &
\includegraphics[width=0.12\linewidth]{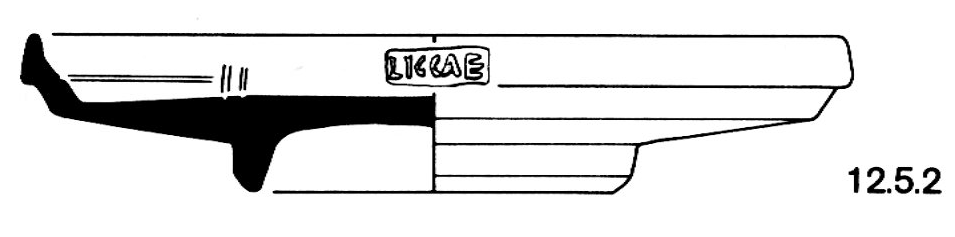} &
\includegraphics[width=0.12\linewidth]{14_1_5.png} \\
\hline
\includegraphics[width=0.12\linewidth]{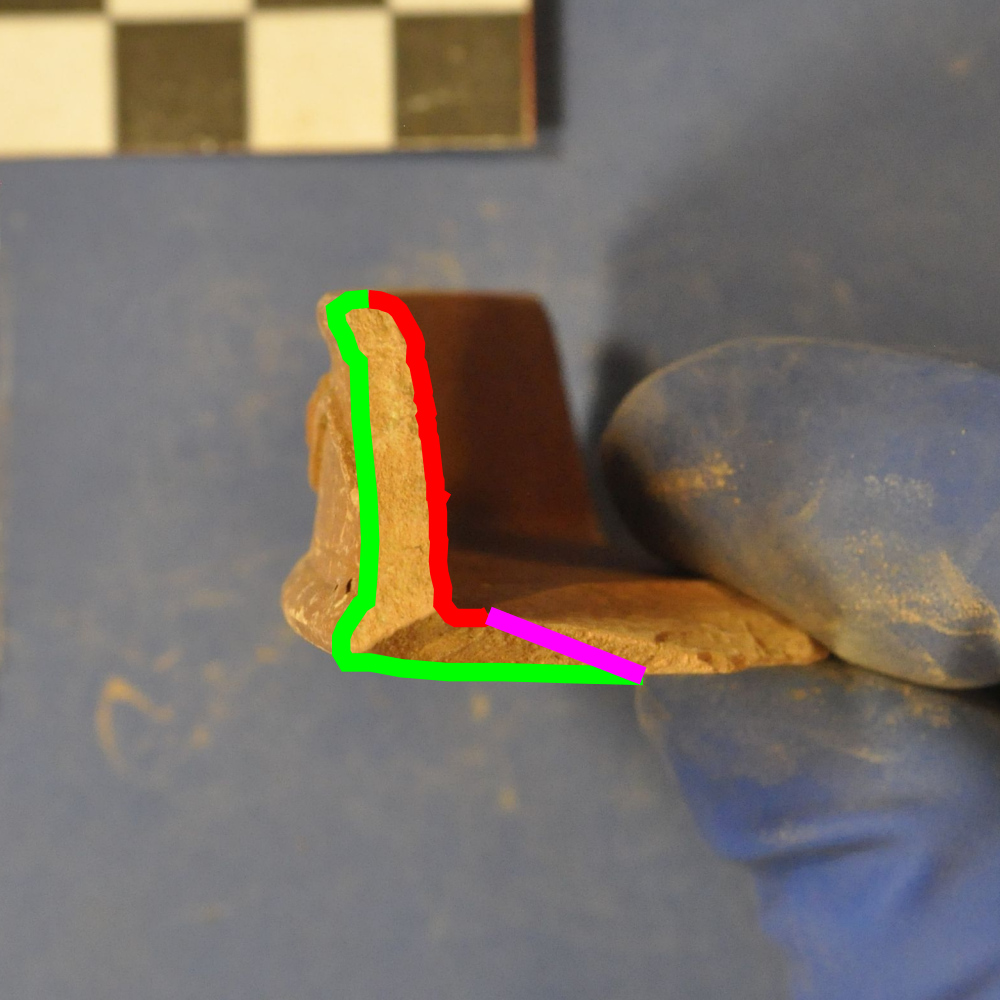} & \includegraphics[width=0.12\linewidth]{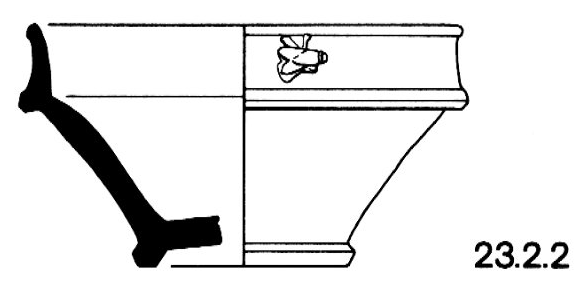} &
\raisebox{1.3cm}{~~~~~~~~(*)\hspace{-1.33cm}} \includegraphics[width=0.12\linewidth]{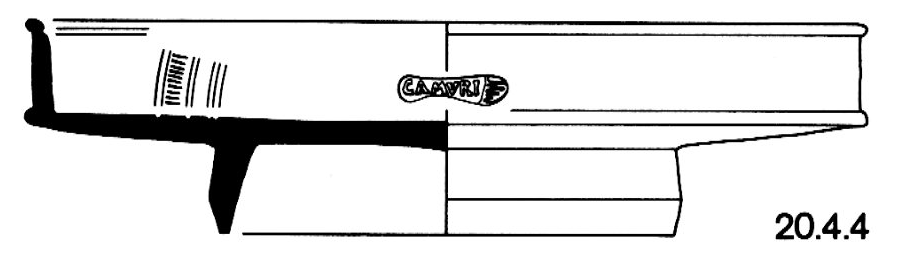} &
\includegraphics[width=0.12\linewidth]{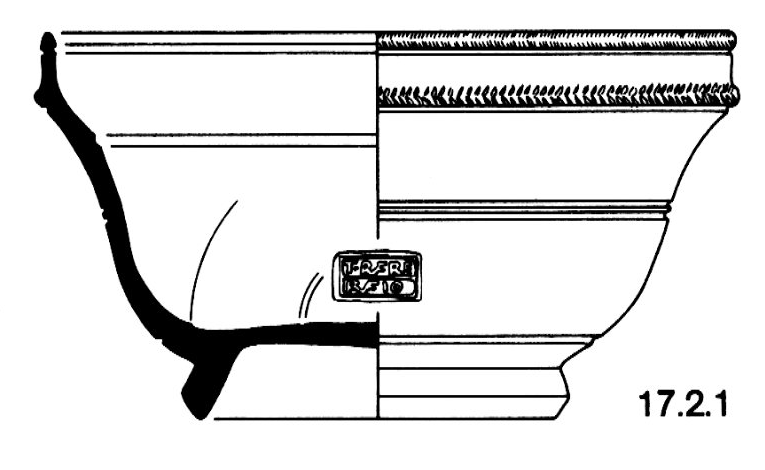} &
\includegraphics[width=0.12\linewidth]{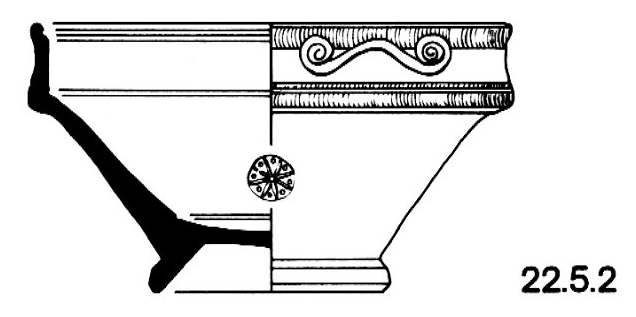} &
\includegraphics[width=0.12\linewidth]{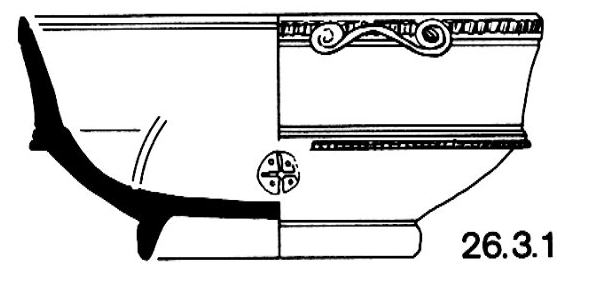} &
\includegraphics[width=0.12\linewidth]{20_4_4.png} \\
\hline
\includegraphics[width=0.12\linewidth]{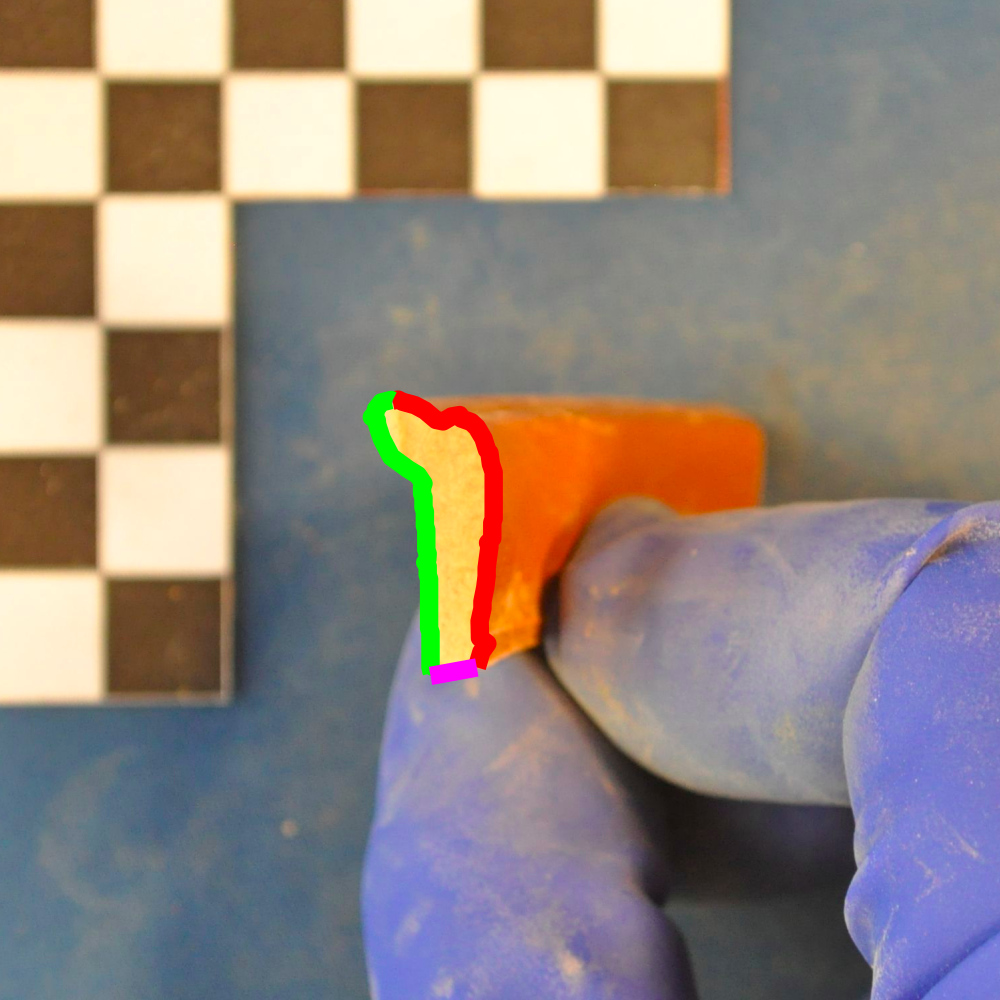} &
\includegraphics[width=0.12\linewidth]{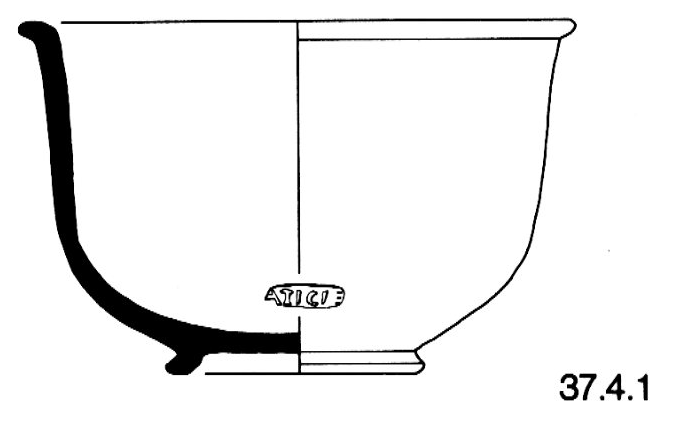} &
\includegraphics[width=0.12\linewidth]{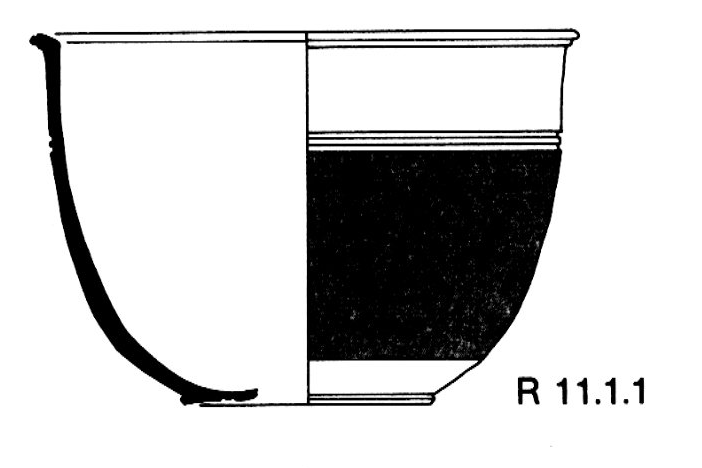} &
\includegraphics[width=0.12\linewidth]{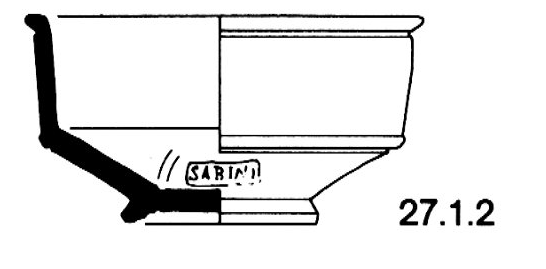} &
\includegraphics[width=0.12\linewidth]{13_3_1.png} &
\includegraphics[width=0.12\linewidth]{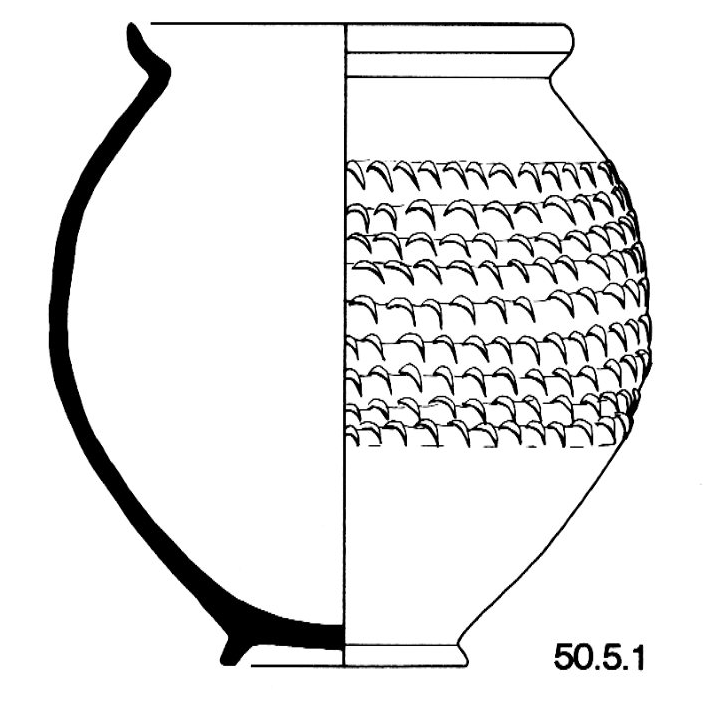} &
\includegraphics[width=0.12\linewidth]{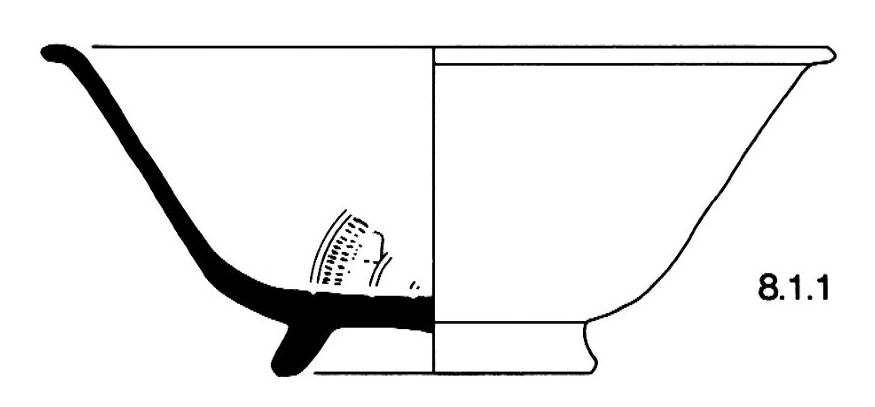}
\\
\hline
\end{tabular}}
\caption{Example sherds and their top-5 results with our model; (*) marks the correct class.}
\label{fig:visual-ranking}
\end{figure*}

The real-world test dataset contains 240 extracted outlines from 29 different top-level classes. Nevertheless, we train our classifier on all 65 classes. Since  the real-world test set is unbalanced, we report  mean accuracy across classes. We also report results on a synthetic test set, which is obtained without the augmentation we apply to the training set, making it ``easier'' in this sense than the training data. Unless otherwise mentioned, all runs use our CareLoss function. 

Table~\ref{tab:cancercorr} reports the results of the various experiments. As may be noted, our OutlineNet's real-world top-2 classification rate is 1.5 times the top-1 classification rate. This indicates that the classes are easily confused, as can be seen in Figure~\ref{fig:visual-ranking}.

We compare our OutlineNet to various baselines. PointNet~\cite{qi2017pointnet} and  PointCNN~\cite{hua2018pointwise} results are given for the 8D feature vector $(x_{\textit{in}}, y_{\textit{in}}, \sin \theta_{\textit{in}}, \cos \theta_{\textit{in}}, x_{\textit{out}}, y_{\textit{out}}, \sin \theta_{\textit{out}}, \cos \theta_{\textit{out}})$ described in Section~\ref{sec:network}, or to the 2D points, as these methods were originally conceived.  When applied to 8D inputs, we enlarge the capacity of these methods, and use the same number of parameters for PointNet as for OutlineNet. 

In the first set of comparisons, CareLoss is used for all methods. As can be seen, PointNet does well on the synthetic data. However, it is not competitive to OutlineNet on the real-world data when using the 8D features. With the 2D points, PointNet is slightly better in the top-1 accuracy than OutlineNet, but not on the other top-$k$ accuracies. PointCNN is not competitive in our experiments, 
showing lower training accuracies than other methods, and a complete failure in generalization to both the test and the real world data. 

PointCNN implicitly requires normalized data, and we therefore retrained it with normalized data, where the sherd outline radius is scaled to fit the unit circle. As shown in the table, this did not provide any significant improvement. Another possible factor for the failure of PointCNN is the fixed sample counts that  it requires. The latter prevent the application of the adaptation to sampling resolution we present in Section~\ref{sec:doc2sherd}.  

These experiments are repeated without the CareLoss. As can be seen, the results for our OutlineNet are reduced along all real-world measurements, excluding the top-2 result.  A similar effect is seen for PointNet on the real-world test data for both the 2D and the 8D configuration. For PointCNN, which is performing almost at random for the real-world data, the results are similar without CareLoss and are omitted. 

To further study the effect of CareLoss, we also compare it to variants where either $u$ or $v$ was set to one (so there is only one weight) and to focal loss reweighting~\cite{lin2018focal} with the recommended parameter of $\gamma=2$. The results show that the dual weighting is important for the real-world top-$k$ results, and especially for the top-5 and top-10 ones. Focal Loss is consistently ineffective in our problem, and it also does not seem beneficial to combine it with CareLoss.

Another group of ablation experiments tests architectural modification: (i) ``OutlineNet w/o separation of in/out (inner and outer outlines)''  skips the group hot encoding of points and angles (effectively discarding the group information); (ii) ``OutlineNet w/o angle information'' denotes dropping the pipeline of processing angles; (iii) ``OutlineNet w/o group-hot encoding'' appends the group id as a one-hot vector of size 2; and (iv) ``OutlineNet w/ a single pipeline'' is similar to PointNet working on 8D vectors but also incorporates the adaptive sampling. As can be seen, all those modifications add to the overall performance when considering top-$k$ for $k>1$. 

This is consistent with our finding for the 2D PointNet: the OutlineNet architecture presents its largest advantage after the top result. This befits the use as a reference tool for domain experts who would be happy to consider a short list of results, as part of the mandatory expert verification, but would be discouraged to use a tool that often completely omits the correct result.

A similar phenomenon can be observed when training without augmentation. In this case, as expected, the performance on the synthetic data is increased. The performance for top-1 and top-2 results on the real data is at least as good. However, going beyond $k=2$, the advantage of augmentation becomes very clear. A plausible reason is that augmentation helps less with the samples that are carefully collected and informative, than it helps with the lower-quality ones.

\begin{figure}
\centering
\includegraphics[width=.7\columnwidth]{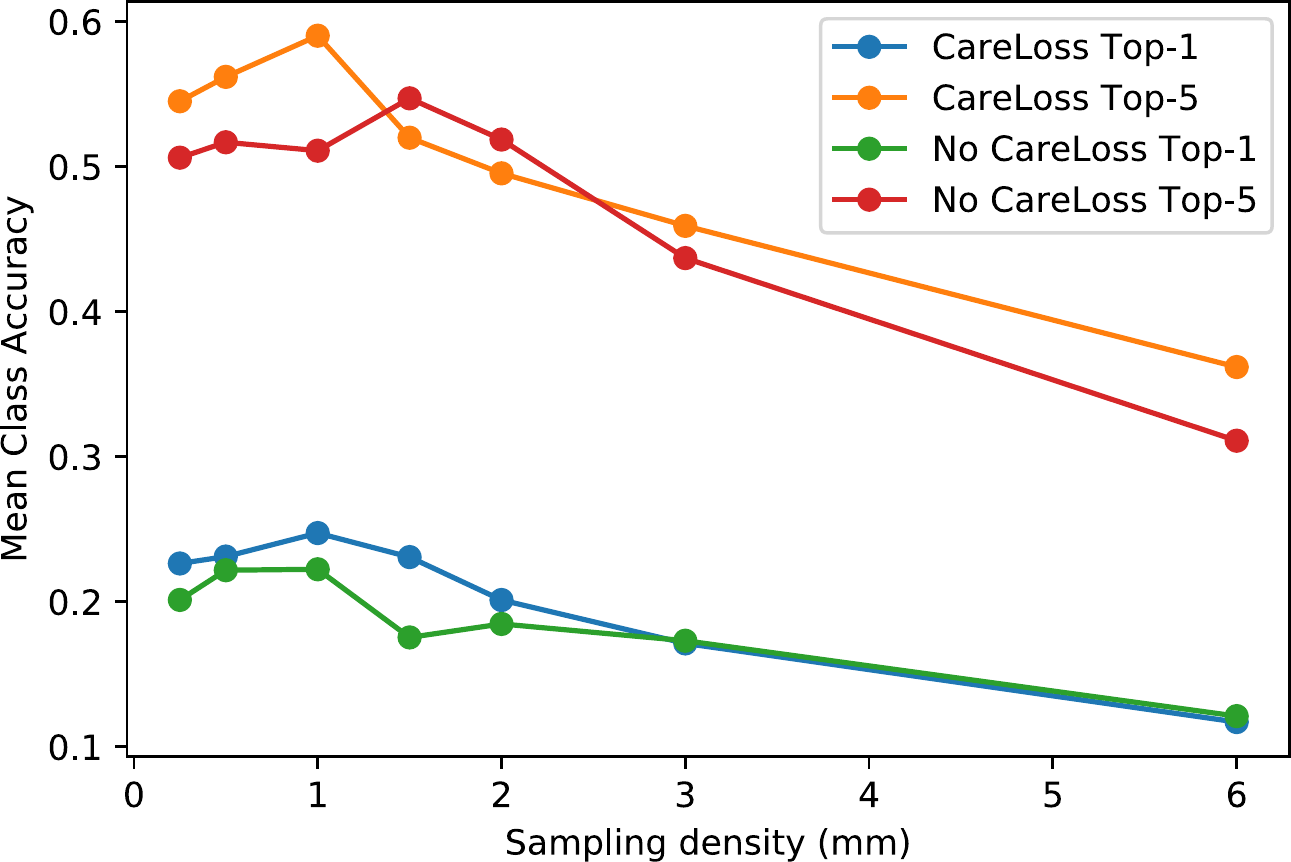}
\caption{Accuracy vs.\@ testing sampling density.}
\label{fig:res}
\end{figure}

The benefit of sampling the test data at a resolution higher than the one used for training is presented in Figure~\ref{fig:res}. The best results for both top-1 and top-5 classification are obtained for a resolution that is double the training resolution of 2mm. 

In addition to the mean performance, we also observe the standard deviation (SD) between classes of the performance, to test if the overall mean success comes at the expense of classes whose performance is left behind. These results are reported in Table~\ref{tab:sd}. As expected, training without augmentation reduces the SD on the synthetic data, but not on the real data.  OutlineNet without the CareLoss, the PointNet 8D and the focal loss methods enjoys a relatively low SD on the top-1 result, but these methods have a relatively low performance there. OutlineNet presents a better SD for top-1 than the other effective methods. In the top-2 ranking, the only methods to show less variance between  classes than OutlineNet are the ones using Focal Loss with OutlineNet. However, these methods are not competitive in their mean performance. Finally,  in the top-5 and top-10 measures, the SD of our method is better than all other methods, with the exception of the PointNet 2D variants at top-5, which are not competitive in this measure. To summarize: no method with a relatively good performance in a top-$k$ measure also displays larger equality among classes than OutlineNet with CareLoss.

We set the parameters of the CareLoss early during development, before the architecture was finalized. However, an analysis of the method's stability to its parameters shows that it is stable and that the results are similar for a wide range of parameter values.

\subsection{Phase II Datasets}

Following the development of the method on the Terra Sigillata Italica (TSI) dataset detailed above, we received three additional datasets. The first was an additional test dataset, also of TSI, collected with the aid of the app. It includes the outlines of a further 96 actual sherds not included in our previous dataset of real data and belonging to 11 classes previously unseen during test. 

Two additional datasets, Terra Sigillata Hispanica (TSH) and Terra Sigillata South Gaulish (TSSG), were handed to us less than 2 weeks and just 3 days, respectively, prior to the submission deadline. These datasets also describe Terra Sigillata pottery, yet from different geographical origins and manufacturers, and belong  to a different set of classes with different typology. (There is no intersection in classes between TSI, TSH, and TSSG.)

Table~\ref{tab:accDatasets2} gives our model's results on the new TSI test set, without performing any retraining or adaptation, using the same complete OutlineNet model from the above experiments. The  accuracy values obtained are even better than the statistics for the prior real-world test set, which was employed in the Phase I experiments reported above. This, therefore, further supports the claim of robustness for our methodology and its applicability as an actual reference tool for archeologists.

Table~\ref{tab:accDatasets2} also reports the results on the two additional datasets using our complete method (OutlineNet with CareLoss and data augmentation), using the same methodology and parameters. No tweaking whatsoever was performed for these datasets on any part of the training and classification process. As can be seen, the pipeline generalize well to \emph{TSH}. While it also succeeds in learning for \emph{TSSG} with similar train/test results on the synthetic data, these evaluation results are a bit lower than for the other datasets. It seems that in many of the outlines of the TSSG dataset, the inner and the outer labels are not correctly set, and due to the short time before the deadline in which this dataset was provided, we have not been able to correct for this yet.

\begin{table*}
\centering
\caption{Mean classification accuracy across all classes for OutlineNet on all {\color{black}four } datasets.}
\label{tab:accDatasets2}
\resizebox{.94687425\textwidth}{!}{
\begin{tabular}{lcccccc}
\hline
& \multicolumn{2}{c}{Synthetic data}& \multicolumn{4}{c}{Real-world test data}\\
\cline{2-3}
\cline{4-7}
Dataset	& Train & Test & Top-1 & Top-2 & Top-5 & Top-10\\
\hline
TSI (Phase I) -- 65 classes (29 in eval), 240 samples & 60.9\% & 70.0\% & 22.0\% & 32.7\% & 57.9\% & 73.7\% \\
TSI (Phase II, only test) -- 65 classes (11 in eval), 96 samples & --\texttt{"}-- & --\texttt{"}--  & 30.5\% & 43.6\% & 62.8\% & 81.3\% \\
TSH -- 98 classes (24 in eval), 218 samples & 60.3\% & 78.6\% & 27.6\% & 40.6\% & 58.4\% & 68.1\% \\
TSSG -- 94 classes (34 in eval), 185 samples  & 57.7\% & 76.6\% & 14.5\% & 25.0\% & 41.9\% & 59.9\% \\
\hline
\end{tabular}
}
\end{table*}

\subsection{Final Results}

The evaluation of the decoration identification method was done on both the mobile and desktop versions, including testing of different lighting conditions (as these were a key factor in the classification results for the first version). The results for the classification are reported below:

\begin{center}
\begin{tabular}{ l c c }
\hline
 & Mobile performance & Desktop Performance \\ 
\hline
Top-1 Accuracy & 55.2\% & 51.0\% \\  
Top-5 Accuracy & 83.8\% & 77.2\% \\
\hline
\end{tabular}
\end{center}

The above analysis was conducted on 49 different genres (out of \emph{84 different genres}), with more than 700 images taken on mobile devices (phones and tablets), and more than 120 images taken with a camera and classified in the desktop app.

Further results show that the accuracy of appearance-based recognition, on both mobile devices and desktop, is not related to the light type, being approximately equal with artificial and natural light.

\section{Discussion}
\label{sec:limitations}

Archeological classification is not made purely based on the shape of the fracture. Additional domain expertise, which is not currently captured in our scheme, enables the archeologist to filter out some classes based on the location of the findings, other findings in the excavation site, and various other considerations. This by itself is not a technological limitation, as this sort of filtering can be implemented on top of the class ranking predicted by our reference tool. However, it means that the gap in the ability to distinguish potsherds based on their shape, vs.\@ human archeologists, is probably much lower than the error rates of our method. 

Another reason to believe that the error rates are probably inflated is that the labelings of  individual potsherds are gathered from accepted labelings that are documented in catalogs and in established collections.  However, in some cases, the exact provenance of the assignment has been lost, and the ground truth classification is likely to contain mistakes.%

To tackle a real-world cross-modality matching problem that presents a large set of compounding challenges,  we have conceived of multiple innovations, including the design of novel data generation techniques, a new shape representation scheme, and a new reweighting method. Our work also provides---beyond multiple technical novelties and a working application---a case study of deep learning applied  to real-world data in a situation where most of the conventional assumptions are grossly violated and the reality gap (the sim2real domain shift) is wide and the simulation must be done with significant care.

The method described in this paper is already deployed in the field as the main part of an archeological reference tool. 
The source code 
and models will shortly be made publicly available as open source. The data is in the process of release to the public.

\subsection*{Acknowledgements}
This research was supported by EU Horizon 2020 grant agreement No.\@ 693548. We thank all the members of the ArchAIDE (archaide.eu) team.

\bibliographystyle{plain}
\bibliography{arch}

\end{document}